\documentclass[letterpaper, 10 pt, conference]{ieeeconf}  
\usepackage{lipsum} 
\usepackage{lscape}
\usepackage{subcaption}
\usepackage{algorithm}
\usepackage{xspace}
\usepackage{physics}
\usepackage{mathtools}
\usepackage{bbold}
\usepackage[noend]{algpseudocode}
\usepackage{amssymb}
\usepackage{booktabs}
\usepackage{adjustbox}
\usepackage{multirow}
\usepackage{wrapfig}
\usepackage{color}
\usepackage{cite}
\usepackage{hyperref}
\usepackage{makecell}

\IEEEoverridecommandlockouts                              

\title{\LARGE \bf
Dynamics as Prompts: In-Context Learning for Sim-to-Real System Identifications
}

\author{Xilun Zhang$^{*1}$, Shiqi Liu$^{*1}$, Peide Huang$^{1}$, William Jongwon Han$^{1}$, Yiqi Lyu$^{1}$, Mengdi Xu$^{1}$, Ding Zhao$^{1}$
\thanks{* indicates Equal Contribution}
\thanks{$^{1}$with Carnegie Mellon University, Pittsburgh, USA
        {\tt\{xilunz, shiqiliu, peideh, wjhan, yiqilyu,  mengdixu, dingzhao\}@andrew.cmu.edu}}%
}

\begin{document}

\maketitle
\thispagestyle{empty}
\pagestyle{empty}

\newcommand{\xilun}[1]{\textcolor{blue}{\textbf{(Xilun: }#1)}}
\newcommand{\shiqi}[1]{\textcolor{cyan}{\textbf{(Shiqi: }#1)}}
\newcommand{\william}[1]{\textcolor{orange}{\textbf{(William: }#1)}}
\newcommand{\abbv}{{CAPTURE}\xspace}

\begin{abstract}
Sim-to-real transfer remains a significant challenge in robotics due to the discrepancies between simulated and real-world dynamics. Traditional methods like Domain Randomization often fail to capture fine-grained dynamics, limiting their effectiveness for precise control tasks. In this work, we propose a novel approach that dynamically adjusts simulation environment parameters online using in-context learning. By leveraging past interaction histories as context, our method adapts the simulation environment dynamics to real-world dynamics without requiring gradient updates, resulting in faster and more accurate alignment between simulated and real-world performance. We validate our approach across two tasks: object scooping and table air hockey. 
In the sim-to-sim evaluations, our method significantly outperforms the baselines on environment parameter estimation by 80\% and 42\% in the object scooping and table air hockey setups, respectively. Furthermore, our method achieves at least 70\% success rate in sim-to-real transfer on object scooping across three different objects. By incorporating historical interaction data, our approach delivers efficient and smooth system identification, advancing the deployment of robots in dynamic real-world scenarios. Demos are available on our project page: \href{https://sim2real-capture.github.io/}{https://sim2real-capture.github.io/}

\end{abstract}

\section{Introduction}  
Learning-based methods like deep Reinforcement Learning (RL) allow robots to tackle complex tasks in areas such as object manipulation~\cite{8460528,lin2024generalize} and locomotion for quadrupedal robots~\cite{10611442,kumar2021rma} and humanoids~\cite{chen2024learning, zhang2024wococo}. 
However, RL's high sample complexity and risks of unsafe exploration~\cite{xu2022trustworthy,wang2023guardians,yao2024constraint} make it necessary to train policies in simulations and then deploy in the real world. 
A key challenge is the sim-to-real gap, caused by discrepancies between simulated and real-world dynamics~\cite{hu2024learning, torne2024reconciling, huang2022robust}, which can lead to catastrophic failures during deployment.

\begin{figure*}[tbp]
    \centering
    \includegraphics[width=\linewidth]{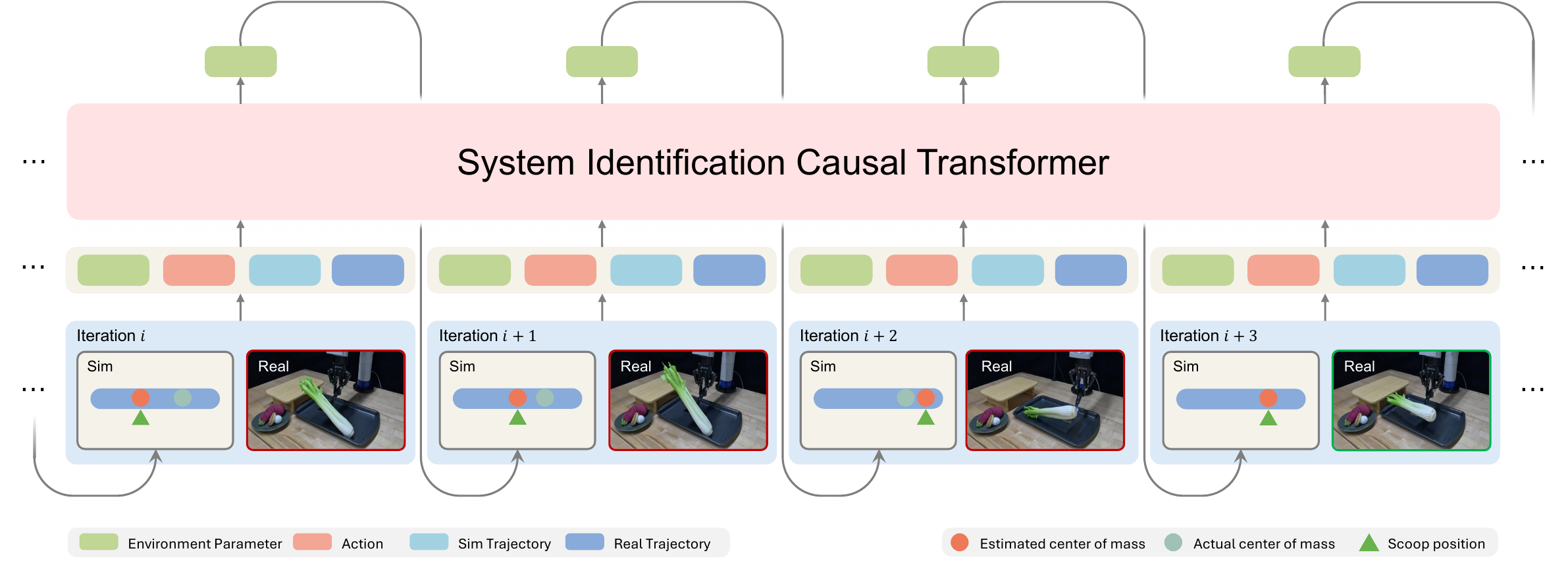}
    \caption{\abbv aims to take the history information to predict the next step environment parameters. SysID causal transformer adapts the simulation environment parameters to match the real-world performance on the fly via next-token prediction. \abbv takes three iterations to identify the correct center of mass of celery.}
    \label{fig:train_pipeline}
\end{figure*}

Traditional sim-to-real approaches aim to develop robust policies by randomizing environment parameters during training, known as Domain Randomization (DR)~\cite{8460528, pmlr-v100-mehta20a}. 
While effective in some cases~\cite{pmlr-v100-mehta20a, 10611442}, DR captures only average dynamics, limiting precision in fine-grained control tasks.
In contrast, System Identification (SysID) methods aim to align the simulation and real-world performance through actively adjusting the simulation environment parameters, which often requiring iterative SysID model updates to test new parameters~\cite{ramos2019bayessim, pmlr-v229-huang23c}. 
For instance, in a kitchen environment, when a robot tries to scoop grilled celery from a pan (Figure~\ref{fig:train_pipeline}), traditional offline SysID methods would involve learning a new SysID model that predict the center of mass of the celery at each iteration, making the process time-consuming and inefficient. 
Humans, on the other hand, can quickly adapt online. A more intuitive solution is to develop a model with online SysID, allowing for more efficient parameter estimation across different environment dynamics.

In-context learning has gained traction as a method for adjusting model behavior without gradient updates, widely used in Natural Language Processing (NLP)~\cite{dong2022survey} and recently applied in robotics to improve generalization~\cite{laskin2022context, grigsbyamago, xu2022prompting}. 
For example, Xu et al.~\cite{xu2022prompting} enhanced the Decision Transformer (DT)~\cite{chen2021decision} by using new task demonstrations as prompts for online adaptation.
Most current in-context learning approaches focus on adapting policies when rewards or expert demonstrations changes, assuming fixed environment dynamics~\cite{grigsbyamago, fu2024context}.
While different dynamics could be framed as diverse tasks in a multi-task RL setting, more than it's a counter-intuitive setting, it also becomes impractical with a high-dimensional continuous environment parameter space, requiring many tasks to capture the full range of behaviors.
In this paper, we explore a novel question: ``\textit{Can we adapt simulation environment parameters using the in-context learning paradigm?}" 
Our goal is to eliminate the optimization loop in SysID, in order to accelerate the parameter estimation process by incorporating the in-context learning ability of transformer models. 

We introduce in-\textbf{C}ontext \textbf{A}da\textbf{PT}ation mod\textbf{U}le for sim-to-\textbf{RE}al system identification, or \abbv, to bridge the sim-to-real gap. 
\abbv aims to dynamically adjust the environment parameters online to align simulated and real-world trajectories using next-token prediction based on past interaction data, which includes simulated trajectories, actions, environment parameters, and real-world trajectories as shown in Figure~\ref{fig:train_pipeline}.  
\abbv frames the SysID problem as an in-context learning formulation, treating interaction histories as ``\textit{context}."
Unlike existing techniques~\cite{kumar2021rma, kumar2022adapting, pmlr-v229-ren23b} that rely on short state-action history, \abbv aims to learn the complex SysID search process itself through rich and multi-episodic interaction history data.
Beyond learning single-step expert parameter matching behaviors, longer interaction histories enables the learned SysID causal transformer to capture a better dynamics representation of the environments.
By incorporating in-context learning, \abbv provides a smoother and more accurate prediction of subsequent environmental parameters and dynamic behaviors.\looseness=-1  

In summary, this study makes the following contributions: 
\begin{enumerate}  
    \item We propose a novel method that can identify real-world environment parameters without any network parameter updates using in-context learning. 
    \item \abbv distills the SysID parameter update process using multi-episode history, rather than relying on a single-step behavior-to-parameter mapping. This approach allows the SysID causal transformer to learn more comprehensive dynamics properties through interactions, which baseline methods struggle to capture.
    \item We evaluate \abbv in two experiments, object scooping and table air hockey, where we report substantial performance increases in both sim-to-sim transfer and sim-to-real transfer.
    \end{enumerate}  

    \begin{figure*}[htbp]
    \centering
    \includegraphics[width=0.99\textwidth]{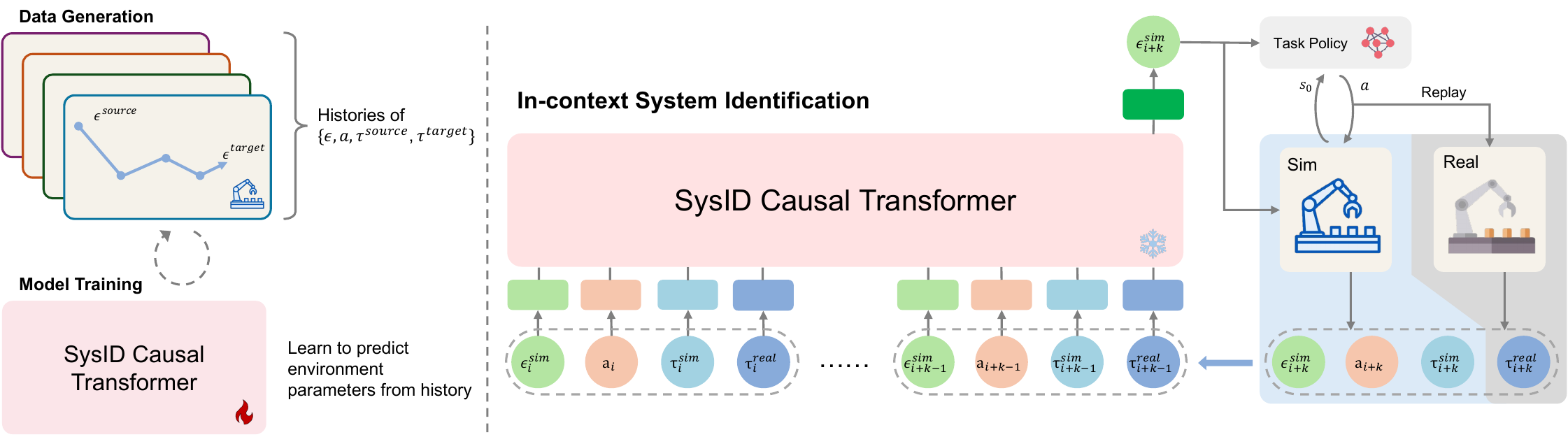}
    \caption{System overview: training and inference pipeline. The SysID causal transformer is trained with multi-episodic parameter update histories. During the in-context SysID, it will take the interaction history as context, and iteratively update the environment parameters online through a task policy rollout in both simulation and the real world.  
    The SysID causal transformer will maintain a fixed-length context window, where in our setting, the length is 4.}
    \label{fig:evaluation}
\end{figure*}
\section{Related Work}

Sim-to-real transfer is a pivotal area of robotics research, focusing on the application of simulation-trained models to real-world tasks. 
DR involves injecting variability into the parameters of the simulation environment regarding dynamical or visual attributes~\cite{8460528, pmlr-v100-mehta20a}, but struggle with over-conservative or average task behaviours.
In the following subsections, prior works on SysID for domain adaptation and in-context learning will be discussed in more detail.

\subsection{Sim-to-Real SysID for Domain Adaptation}

There are two primary approaches on SysID for sim-to-real transfer: offline and online. 
Offline SysID typically requires iterative refinement of the identification module through repeated training cycles\cite{pmlr-v229-huang23c, murooka2021exi, chebotar2019closing, ramos2019bayessim, muratore2022neural, lim2022real2sim2real}.  
In contrast, online SysID focuses on the determination of environment parameters or latent variables without the need for model updates. 
This approach has proven effective in highly dynamic systems, employing strategies such as RMA~\cite{kumar2021rma}, which leverages short-term historical state-action pairs to infer environment dynamics~\cite{yu2017preparing, evans2022context, kumar2022adapting, allevato2020tunenet}. 
\cite{memmelasid} describes exploring the object dynamics through curiosity-driven exploration first and then deploying on the task environment.\cite{pmlr-v229-ren23b} propose a meta-learning framework, prioritizing task-specific adaptation over simple trajectory alignment. 
In addition to aligning environment parameters, \cite{jiang2024transic} introduced a human-in-the-loop correction method to mitigate the sim-to-real gap. More recently, \cite{dai2024acdc} proposed reconstructing real-world environmental variations in simulation to enhance the generalizability on real-world policy deployment.
Most relevant to our work, IIDA~\cite{evans2022context} uses long-term historical state action pairs to infer latent real-world dynamic models. In contrast, our method focuses on distilling the sim-to-real parameter update process to create more accurate simulation environments, effectively closing the sim-to-real gap.

\subsection{In-context Learning in Robotics}
In-context learning has garnered significant attention in NLP~\cite{dong2022survey, krishnamurthy2024can} and computer vision~\cite{wang2023images} due to its remarkable ability to infer tasks from context. 
This ability to infer tasks through contextual information, such as expert demonstrations, allows for adaptation to new tasks without updating the model's weights~\cite{min2022rethinking}, which has been shown to be beneficial in robotics settings~\cite{xu2022prompting, xu2023hyper, zhu2024incoro, di2024keypoint,yu2024fewshotincontextpreferencelearning, jiang2023learning}. 
The potential of in-context learning for generalizing to unseen tasks has been further explored in recent studies. 
Laskin et al.~\cite{laskin2022context} employed transformer models to distill the RL learning history, showing RL algorithms can be distilled into transformer models and successfully in-context adapt to new goal settings~\cite{grigsbyamago}.
Previous work on in-context adaptation has either focused on RL algorithm distillation or policy generalization abilities, where \abbv focuses on learning environment parameters through interaction histories.
\section{Methodology}

Rather than directly adapting the task policy, we prioritize leveraging historical data—including past environment parameters, task state trajectories, and task actions—to estimate next-iteration environment parameters. 
This approach aims to align simulation dynamics with real-world performance.
We assume that as the discrepancy between simulation and real-world environment parameters decreases, the sim-to-real performance gap will naturally narrow. 
This process is guided by the underlying monotonic properties of the environment parameter adjustments.
We start with a description of the problem formulation in Section~\ref{Problem Formulation}. 
Following with three key modules in our pipeline: Section~\ref{environment conditioned RL} describes the task policy training, Section~\ref{data generation} describes how we generate efficient source-to-target adaptation iterations, and Section~\ref{SysID causal transformer} defines different components in the SysID causal transformer structure. 
The main components of \abbv pipeline is demonstrated in Figure~\ref{fig:evaluation}, where it consists the data generation, model training, and inference pipeline.

\subsection{Problem Formulation}
\label{Problem Formulation}

In this section, we define the objective of our approach: aligning simulation and real-world dynamics performance through predicting accurate real-world environment parameters.
We begin by introducing the simulation parameters, followed by the task policy and data generation notations, and conclude with the SysID causal transformer notations for domain adaptation.

\textbf{Environment Parameter Space.} We define the task-related environment parameter space $\epsilon \in \mathcal{E}$, which encapsulates different environment parameter values such as the center of mass and sliding friction. We also assume that the tunable environment parameter space $\mathcal{E}$ in simulation is finite and bounded, encompassing properties of different objects. We modify the environment parameters with Robosuite~\cite{zhu2020robosuite}, which provides API for modifying the environment parameters through Python code.\looseness=-1

\textbf{SysID Causal Transformer and Interaction Histories.} During the SysID causal transformer and data collection, we treat previous SysID iterations as context, including simulated state trajectories $\tau^{sim} = \{s^{sim}_{0},s^{sim}_{1}, \dots,s^{sim}_{T}\}$, real state trajectories $\tau^{real} = \{s^{real}_{0},s^{real}_{1}, \dots,s^{real}_{T}\}$, rollout action $a \sim \pi(a|s_{0}, \epsilon)$, and the past environment parameters $\epsilon$. A robust SysID process must be capable of exploring complex parameter behaviors. To achieve this, we leverage a causal transformer designed to infer environment parameters based on historical interaction data. Following \cite{laskin2022context}, we treat these sequential interactions as \textit{history}, where current environment parameters depend on previous SysID iterations. Formally, we define the \textit{history} as:

\begin{equation}
h_{i} := \left(\epsilon^{sim}_{i-k}, a_{i-k}, \tau^{sim}_{i-k}, \tau^{real}_{i-k}, \ldots, \epsilon^{sim}_{i-1}, a_{i-1}, \tau^{sim}_{i-1}, \tau^{real}_{i-1}\right)
\label{eq:agent_history}
\end{equation}

\noindent where $h_{i}$ is the history containing the past $k$ iterations at $i$-th iteration. 
Our goal is to learn a causal transformer such that it can replicate the SysID process given \textit{history}.
We define the SysID causal transformer, $P_{\theta}$, with the objective of modeling the distribution of simulation parameters conditioned on the history. 
This approach aims to distill the SysID algorithm through interaction histories, such that the simulated trajectories $\tau^{sim}$ aligns with the real-world trajectories $\tau^{real}$, thereby bridging the sim-to-real gap.
The goal of distilling SysID algorithm is to learn the underlying search capabilities from this process by predicting the next iteration in the history.
The optimization objective can be formalized as:

\begin{equation}
\theta^{*} = \arg \min_{\theta} \left[ \mathcal{L}\left( P_\theta(h_{i}), \epsilon^{sim}_i \right) \right]
\label{eq:minimization_hat}
\end{equation}

where $P_\theta(h_{i})$ represents the predicted next environment parameters from the model, $\mathcal{L}(\cdot)$ is the Mean-Square-Error (MSE) loss function that measures the discrepancy between the predicted and the ground-truth next-iteration environment parameters.

\subsection{Environment-Conditioned RL Training}
\label{environment conditioned RL}
The environment-conditioned RL task policy \( \pi(a \mid s_0, \epsilon) \) is trained to adapt to varying environment parameters \( \epsilon \in \mathcal{E} \). 
For each episode, \( \epsilon \) is sampled uniformly from the parameter space \( \mathcal{E} \). 
Within the episode, the agent selects an action \( a \) from \( \pi(a \mid s_0, \epsilon) \), considering the initial state \( s_0 \) and current simulation environment parameter \( \epsilon \).  
This action is executed, producing a state trajectories \( \{s_1, s_2, \ldots, s_T\} \) and a reward $r$.
Each episode $\{a, r, s_0, \epsilon\}$, is stored in the replay buffer. 
After certain episodes, the policy is updated using Soft Actor-Critic (SAC)~\cite{haarnoja2018soft}, refining actions for smoother domain adaptation with predicted parameters.

\subsection{Source-to-Target SysID Iteration Generation}
\label{data generation}
In the data generation process, we developed source-to-target adaptation transitions that mimic sim-to-real adaptation. 
Each iteration includes four elements: the current simulation environment parameter $\epsilon_{i}$, the rollout action $a_{i}$, the simulated trajectories $\tau^{source}_{i}$, and the collected target environment trajectories $\tau^{target}_{i}$ under the same action $a_{i}$. 
The trajectories and actions are obtained through simulation rollouts using an environment-conditioned task policy.
Although random action can be used here to collect training data, task policy is preferred to ensure sufficient task-related environment interactions.

In simulation, both source and target values are known, allowing for direct single-step mapping from source to target.
However, this approach often performs poorly in real-world deployment when the target's dynamics representation (state trajectories) lacks sufficient detail.
Rather than learning a single-step mapping, we focus on learning a search algorithm that finds the target environment parameter with dynamic representations. 
The duration of the parameter iteration history $L$ indicates the number of iterations that we pre-defined to generate a complete transition sequence from $\epsilon^{source}$ to $\epsilon^{target}$. We pick a transition number $L=7$ during data generation.\looseness=-1  

\begin{figure}[tp]
    \centering
    \includegraphics[width=0.9\linewidth]{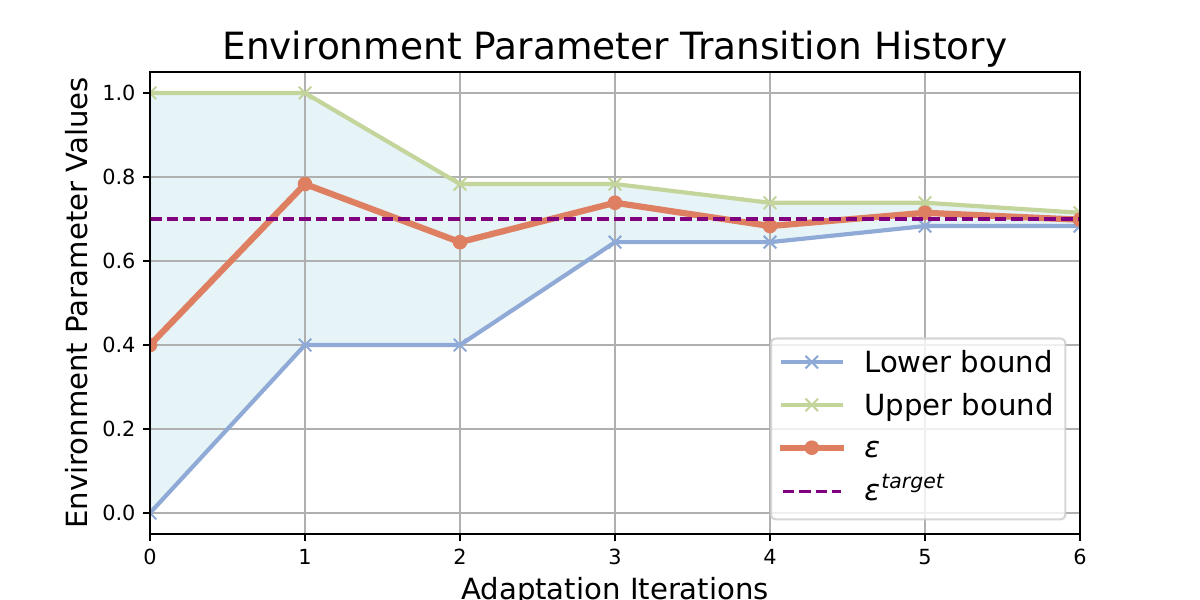}
    \caption{A environment parameter transition history from $\epsilon^{source}$ to $\epsilon^{target}$, with gradually shrank upper and lower bounds of the search space.
    }
    \label{fig:binary_search}
\end{figure}

In the sim-to-real SysID setting, a search algorithm must balance exploration and precision, as it lacks the ground-truth target value and relies only on performance labels (higher or lower). Linear interpolation is suboptimal here because it limits exploration during adaptation. 
To overcome this, we propose emulating a randomized binary search process~\cite{martinez1998randomized}, which optimally navigates a constrained space by dynamically adjusting the upper and lower search bounds at each iteration. 
To further promote exploration, we use a beta distribution when selecting the environment parameters for the next iteration.
An ablation study is discussed in Section~\ref{baselines} on how different search algorithms impact parameter estimation. 
The transition iteration generation process is illustrated in Figure~\ref{fig:binary_search}, and the formal pseudocode is described in Algorithm~\ref{alg:data_collection}.

\begin{algorithm}[tp]
    \caption{Source-to-Target SysID Iteration Generation}
    \label{alg:data_collection}
    \begin{algorithmic}[1]
        \State Initialize data buffer $\mathcal{D}$
        \State Choose parameter transition iteration length $L$
        \State Choose symmetric beta distribution parameter ${\alpha}$
        \For{$n = 1$ to $N$}  \Comment{This loop can be run in parallel}
            \State Sample $\epsilon^{\text{source}}, \epsilon^{\text{target}}$ from space $\mathcal{E}$
            \State Let $l$ be the dynamic lower bound of space $\mathcal{E}$
            \State Let $u$ be the dynamic upper bound of space $\mathcal{E}$
            \State Set $\epsilon_0 = \epsilon^{\text{source}}$
            \For{$i = 0$ to $L$}
                \State Sample action $a_i \sim \pi(a_{i} \mid s_0, \epsilon_{i})$
                \State $\tau^{\text{source}}_{i} \leftarrow$ rollout in $sim(\epsilon_{i})$ with $a_i$
                \State $\tau^{\text{target}}_{i} \leftarrow$ rollout in $sim(\epsilon^{\text{target}})$ with $a_i$
                \For{$j = 1$ to $\dim(\mathcal{E})$}
                    \If{$\epsilon_i[j] < \epsilon^{\text{target}}[j]$}
                        \State Update lower bound: $l[j] = \epsilon_i[j]$
                    \Else
                        \State Update upper bound: $u[j] = \epsilon_i[j]$
                    \EndIf
                    \State Sample $r$ from $\mathrm {B}(\alpha, \alpha)$ 
                    \State Set $\epsilon_{i+1}[j] = r \left(u[j] - l[j]\right) + l[j]$
                \EndFor 
                \State Store trajectory $h[i] = \{\epsilon_{i}, a_{i}, \tau^{\text{source}}_i, \tau^{\text{target}}_i\}$
            \EndFor 
            \State Update data buffer: $\mathcal{D} \leftarrow \mathcal{D} \cup h$
        \EndFor 
    \end{algorithmic}
\end{algorithm}

\subsection{SysID Causal Transformer}
\label{SysID causal transformer}
To model the sequential relations. We adopted the architectural structure from DT\cite{chen2021decision}, which uses GPT-2\cite{radford2019language} as our transformer backbone.
Given the collected SysID parameter transition histories, $\mathcal{D}$, our goal is to distill the binary search process through parameter transition sequences with length $L$, where each iteration represents an adaptation iteration. 
The model predicts the next environment parameter $\hat{\epsilon}_{i+1}$ at iteration $i$ using a next-token prediction framework with a shifted input setup~\cite{radford2019language}.
We sample a multi-episode window of size $k$ from $\mathcal{D}$, where $k$ is a subsequence of the full $L$ iterations. 
The SysID causal transformer processes this history to predict the next-iteration environment parameter. Each iteration block contains $2 + 2T$ tokens: one action, one parameter, and $T$ state trajectory tokens for both simulated and real rollouts.

During rollout, the model attends to preceding tokens to predict $\epsilon_{i+1}$ using relative timestep embedding~\cite{al2019character} to focus on subsequence order. 
Starting with initial tokens $\{ \epsilon_0, a_{0}, {\tau}^{sim}_{0}, {\tau}^{real}_0 \}$, we update actions with an environment-conditioned policy $\pi$ in the new simulation $\epsilon_{i+1}$ and initial state, obtaining updated trajectories $\tau^{sim}_{i+1}$ and $\tau^{real}_{i+1}$. 
The process is detailed in Algorithm~\ref{alg:main}. 
\begin{algorithm}[t]
   \caption{SysID Causal Transformer Training and Evaluation}
   \label{alg:main}
\begin{algorithmic}[1]
   \State Environment-conditioned task policy $\pi$
   \State Collected SysID transition history buffer $\mathcal{D}$
    \State Initialize SysID causal transformer $P_{\theta}$ 
    \State Initialize SysID causal transformer window size $k$
    \State // \textbf{SysID causal transformer training}
    \While{$P_\theta$ not converged}
        \State Sample multi-episodic $k$ subsequence from $\mathcal{D}$:
        \begin{equation*}
            \begin{aligned}
            h_{i} = &\left( 
            \epsilon^{sim}_{i-k}, a_{i-k}, \tau^{sim}_{i-k}, \tau^{real}_{i-k}, \ldots, \epsilon_{i}, a_{i}, \tau^{sim}_{i}, \tau^{real}_{i} \right)
            \end{aligned}
        \end{equation*}
        \State Calculate shifted input loss $\vert\vert P_{\theta}(h_{i-1}) - \epsilon_{i} \vert\vert_2$
        \State Backpropagate to update $P_{\theta}$
        \EndWhile
    \State // \textbf{In-context SysID with env-conditioned policy}
    \For {$i = 0, \ldots, \textit{MaxIters}$}
        \State $\tau^{sim}_{i} \leftarrow$  rollout $a_{i} \sim \pi(a_{i} \mid s_{0},\epsilon_{i})$ in $sim(\epsilon_{i})$
        \State $\tau^{real}_{i} \leftarrow$  rollout $a_{i}$ in unknown real environment  
        \State Predict $\epsilon_{i+1} = P_{\theta}( \{\epsilon_{x}, a_{x}, \tau^{sim}_{x}, \tau^{real}_{x}\}_{x=i}^{i-k})$
    \EndFor
\end{algorithmic}
\end{algorithm}
\section{Experiments}

We conducted two sets of experiments to evaluate the performance of \abbv: object scooping and table air hockey. 
In both tasks, we demonstrated that \abbv significantly outperforms the baselines in both sim-to-sim and sim-to-real transfer scenarios. 
The experiment setups will be explained in Section~\ref{experimental Setups}, followed by descriptions of our baseline and ablation methods in Section~\ref{baselines}. 
The sim-to-sim evaluation results compared with baselines and ablations results are detailed in Section~\ref{Sim-to-Sim SysID Evaluation}, and the sim-to-real experiment results compared with baselines are described in Section~\ref{Sim-to-Real SysID Evaluation}.\looseness=-1  

\subsection{Experimental Setups} \label{experimental Setups}

We evaluate our algorithm using two tasks: object scooping and table air hockey.
For object scooping, inspired by~\cite{memmelasid, shi2023robocook}, the goal is to identify the object's center of mass in kitchen scenarios, which often involve complex items like celery, carrots, and eggplants with varying centers of mass. 
We aim to determine the balance point for successful scooping through online interactions using \abbv.

\begin{figure*}[tbp]
    \centering
    \begin{subfigure}[b]{0.24\textwidth}
        \centering
        \includegraphics[width=0.95\textwidth]{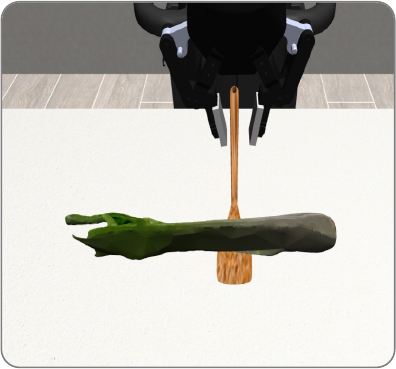}
        \subcaption{Simulated Object Scooping}
        \label{fig:combined_env_1}
    \end{subfigure}
    \hfill
    \begin{subfigure}[b]{0.24\textwidth}
        \centering
        \includegraphics[width=0.95\textwidth]{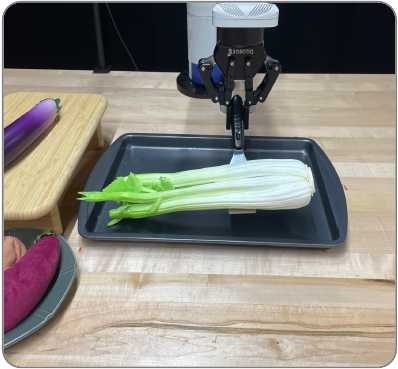}
        \subcaption{Real-world Object Scooping}
        \label{fig:combined_env_2}
    \end{subfigure}
    \hfill
    \begin{subfigure}[b]{0.24\textwidth}
        \centering
        \includegraphics[width=0.95\textwidth]{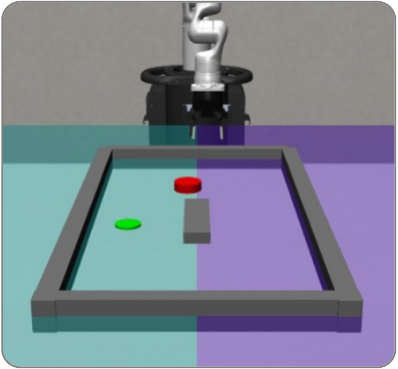}
        \subcaption{Simulated Table Air Hockey}
        \label{fig:combined_env_3}
    \end{subfigure}
    \hfill
    \begin{subfigure}[b]{0.24\textwidth}
        \centering
        \includegraphics[width=0.95\textwidth]{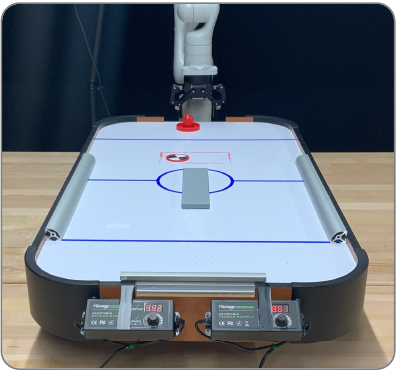}
        \subcaption{Real-world Table Air Hockey}
        \label{fig:combined_env_4}
    \end{subfigure}
    \caption{Experiment setups for both object scooping and table air hockey.}
    \label{fig:combined_env}
\end{figure*}

In table air hockey, we test \abbv with a higher-dimensional parameter and action space~\cite{pmlr-v229-huang23c, chuck2024robot}. 
This task requires precise control and adaptability to match simulated and real-world dynamics. 
Tunable environment parameters are listed in Table~\ref{table:DR_range}, with setups shown in Figure~\ref{fig:combined_env}.

\textbf{Object Scooping.} 
In this task, our objective is to identify the optimal scooping points during food transfer from one toasting pan to another using a spatula. 
In this setting, \abbv needs to identify the center of mass noted as $X_{com}$, and then scoop at the corresponding placement such that the object can be balanced on the spatula. 
The range of $X_{com}$ is defined based on the relative position of the objects, where $-1.0$ means the center of mass located at the most left of the object, and vice versa. 
To handle pose estimation uncertainties, a classifier labels the object as tilted left (-1), right (1), or balanced (0) and uses them as state trajectories.

\textbf{Table Air-Hockey.}  
The setup involves a robot-controlled mallet hitting a puck on an air-hockey table.
The table is divided into left and right sections with different friction levels, causing varied puck behavior.
We expect \abbv to learn surface friction and damping differences from both sides via incorporating context information. 
The five parameters considered are left-surface friction $\mu_{left}$, right-surface friction $\mu_{right}$, left-wall damping $\zeta_{left}$, right-wall damping $\zeta_{right}$, and puck damping $\zeta_{puck}$. 
Lower absolute damping values make objects more responsive, and trajectory evaluation is based on the sum of point-wise L2 distances.

\begin{table}[tbp]
\centering
\caption{Tunable Environment Parameters in Simulation}
\label{tab:air hockey tunable parameters}
\resizebox{\linewidth}{!}{
\begin{tabular}{cccc}
\toprule
\multicolumn{1}{l}{Environment}     & Notion           & Description            & Range            \\ 
\midrule
\multicolumn{1}{l}{Object Scooping} & $X_{com}$        & Object Center of Mass  & {[}-1.0, 1.0{]}  \\ \hline
\multirow{5}{*}{Table Air Hockey}   & $\mu_{left}$     & Table Sliding Friction & {[}0.03, 0.07{]} \\
                                    & $\mu_{right}$    & Table Sliding Friction & {[}0.03, 0.07{]} \\
                                    & $\zeta_{mallet}$ & Mallet Damping         & {[}-15, -3{]}    \\
                                    & $\zeta_{left}$   & Wall Damping           & {[}-40, -3{]}    \\
                                    & $\zeta_{right}$  & Wall Damping           & {[}-40, -3{]}    \\ 
                                    \bottomrule
\end{tabular}
}
\label{table:DR_range}
\end{table}

\subsection{Baselines and Ablations}
\label{baselines}

To discover how different module of \abbv affects the performances, the baselines aim to demonstrate the effectiveness of context history during rollout. The ablations are meant to demonstrate how different data generation methods affect the performance. 
We have compared \abbv with two ablations in sim-to-sim evaluation and three baselines methods in both sim-to-sim and sim-to-real evaluation.

\textbf{Baselines.} \abbv distills the sim-to-real adaptation process to learn an efficient transition from source to target. 
We compare with the following baselines for online adaptation tasks: Expert Distillation (ED)~\cite{laskin2022context}, TuneNet~\cite{allevato2020tunenet}, and DR~\cite{8460528}. ED is similar to \abbv but with expert SysID training data consists of one-iteration source-to-target parameter adaptation, rather than learning histories. To make a fair comparison, we have also implemented the TuneNet~\cite{allevato2020tunenet} algorithm with a transformer backbone, where the model follows the ED setting but with residual parameter updates.

\textbf{Ablations on Different Data Generation Approaches.} 
We modify the data generation module to demonstrate the effectiveness of our distilled searching algorithm over others, including linear interpolation (linterp) and the standard binary search method without randomness (binary), while selecting the next iteration parameters. Linear interpolation randomly sample $L$ points between source and target environment parameters and orderly constructs the SysID transition.
The standard binary search method~\cite{sikorski1982bisection} follows a similar setting as ours. However, it does not consider the random beta distribution, it only selects the middle point between the upper and lower bound.

\subsection{Sim-to-Sim SysID Evaluation}
\label{Sim-to-Sim SysID Evaluation}

In the sim-to-sim transfer, we evaluate whether \abbv can align trajectories by adjusting the environment parameters in-context without updating the model's parameters. 
We simulated 100 pairs of random environment parameters to mimic unknown real dynamics and test the performance across three seeds. 
For each pair, one simulation environment is designated as the ``real" (target) environment, where only the dynamics performance is provided to the model, not the parameters.
To improve parameter estimation independent of actions, we roll out the model with an environment-conditioned policy for online evaluation, as described in Section~\ref{environment conditioned RL}. 
In the results, baseline methods are shown with solid lines, while dashed lines indicate different ablation settings for data collection.

\begin{figure}[t]
    \centering
    \includegraphics[width=\linewidth]{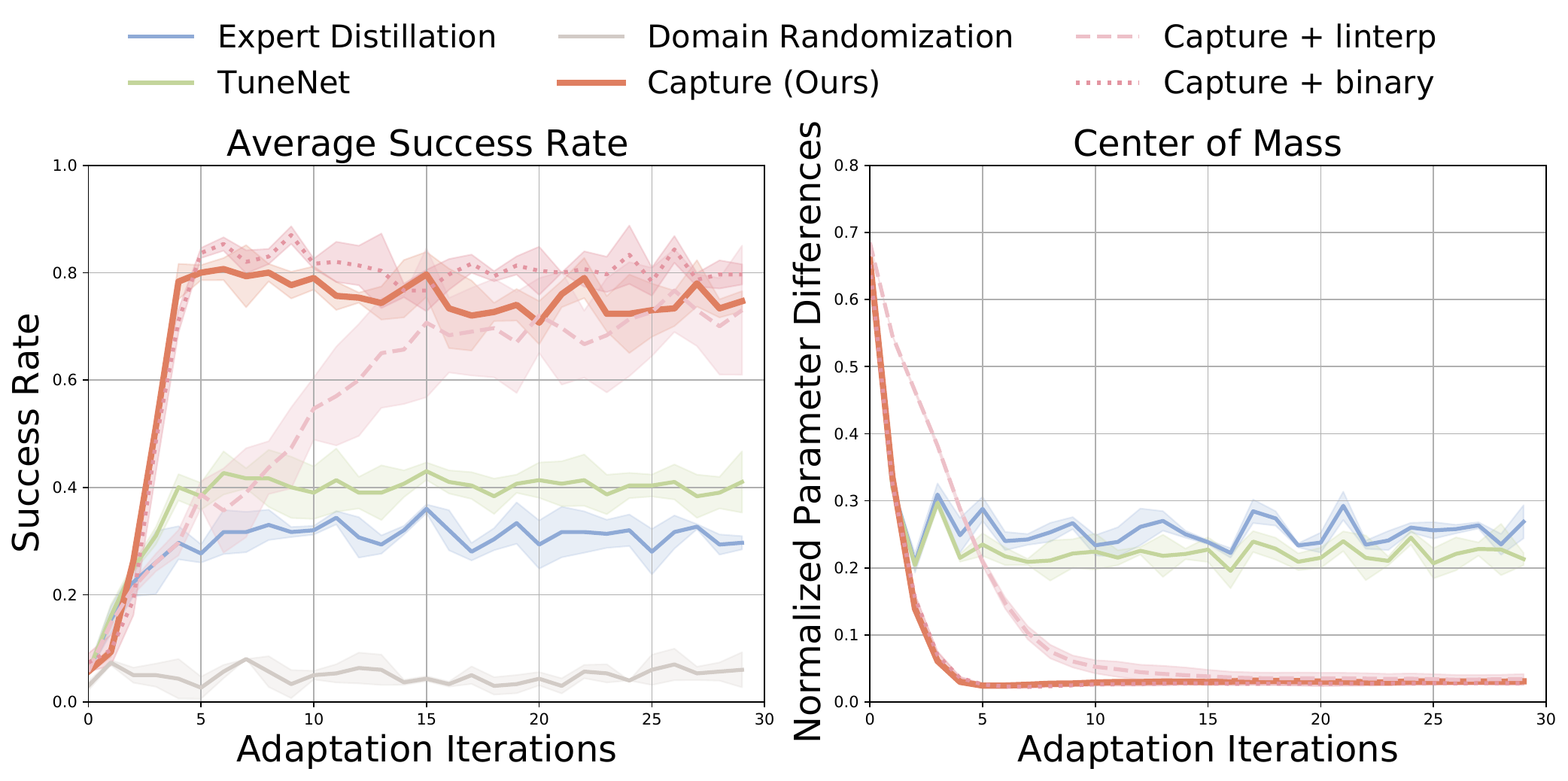}
    \caption{Object scooping sim-to-sim transfer parameter estimation and success rate performance. 
 \abbv identifies objects' center of mass after around 4 iterations. 
 }
    \label{fig:combined_scooping}
\end{figure}

\textbf{Object Scooping Sim-to-Sim Evaluation.} 
In the sim-to-sim transfer, we evaluated the normalized context differences, which are one-dimensional in this setting, as shown in Figure~\ref{fig:combined_scooping}. 
Since we use an angle classifier to reduce dependence on accurate sensor data during real-world deployment, reporting trajectory differences becomes irrelevant, as the trajectory in this context is represented solely by a label.
Instead, we measure the task's success rate, defined as lift the object with label (0).
Figure~\ref{fig:combined_scooping} shows that \abbv achieves a success rate 50\% higher than other SysID methods and 70\% higher than the DR approach. 
This is expected, as the baselines lack historical interaction data, making identification only dependent on current scooping points. 
In contrast, \abbv uses a rich previous interaction history, allowing it to gradually narrow down the center of mass search space.

\begin{figure*}[ht]
    \centering
    \includegraphics[width=\linewidth]{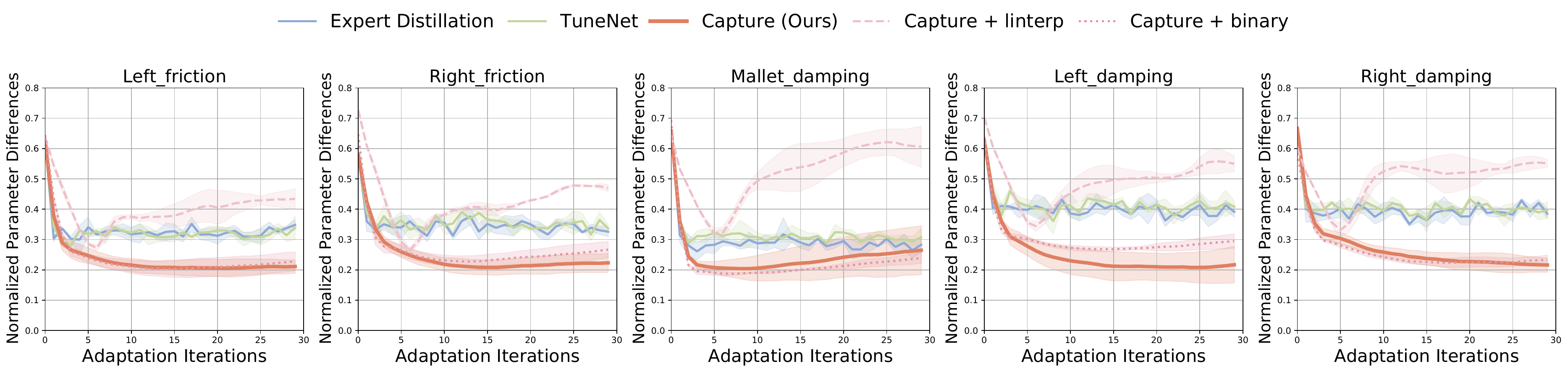}
    \caption{Table air hockey sim-to-sim transfer parameter estimation performances. The red lines represent our proposed method \abbv, which outperforms the baseline methods in all five parameters. Our approach reaches around 0.2 differences after 7 adaptation iterations, where the baselines converge at 0.35 for most parameters.}
    \label{fig:push_one_context_align}
\end{figure*}

\textbf{Table Air Hockey Sim-to-Sim Evaluation.} In Figure~\ref{fig:push_one_context_align}, \abbv offers better parameter estimation with more stable and accurate adaptation curves. In scenarios where environment parameters require rollout histories, baselines struggle due to their inability to account for historical interactions. 
For instance, while the ED method might successfully detect the left wall after hitting it, it tends to forget earlier right wall interactions. This short-term memory leads to faster adaptation in simple environments but falls short in more complex ones. 
In dynamic settings, where SysID needs to identify parameters on both sides for sustained task performance, maintaining a history of parameter updates becomes critical, as it informs subsequent iterations.

\begin{table}[tp]
\centering
\caption{Sim-to-Sim Table Air Hockey Trajectory Differences in Meters over 3 Seeds. \textbf{Bold} font means the best.}
\label{tab:air hockey sim-to-sim traj}
\resizebox{\linewidth}{!}{  
\begin{tabular}{c|ccccc}
\toprule
\multicolumn{1}{c|}{\multirow{2}{*}{Method}} & \multicolumn{5}{c}{Adaptation iterations}                                                                                                 \\  
\multicolumn{1}{l|}{}                        & 5 iterations                   & 10 iterations                  & 15 iterations                  & 20 iterations                  & 30 iterations                  \\ 
\midrule
ED                                           & 0.25$\pm$0.01 & 0.26$\pm$0.03 & 0.27$\pm$0.02 & 0.27$\pm$0.01 & 0.26$\pm$0.01 \\ \hline
DR                                           & 0.34$\pm$0.03 & 0.31$\pm$0.02 & 0.34$\pm$0.04 & 0.31$\pm$0.03 & 0.33$\pm$0.00 \\ \hline
TuneNet                                           & 0.29$\pm$0.02 & 0.27$\pm$0.01 & 0.27$\pm$0.01 & 0.26$\pm$0.01 & 0.25$\pm$0.02 \\ \hline
\abbv + linterp                                  & 0.22$\pm$0.02 & 0.23$\pm$0.01 & 0.26$\pm$0.03 & 0.24$\pm$0.02 & 0.22$\pm$0.01 \\ \hline
\abbv + binary                                   & 0.20$\pm$0.01 & 0.17$\pm$0.01 & 0.16$\pm$0.02 & 0.17$\pm$0.01 & 0.18$\pm$0.01 \\ \hline
\abbv                                           & \textbf{0.20$\pm$0.01} & \textbf{0.16$\pm$0.01} & \textbf{0.14$\pm$0.01} & \textbf{0.14$\pm$0.01} & \textbf{0.15$\pm$0.01} \\ 
\bottomrule
\end{tabular}
}
\end{table}

In Table~\ref{tab:air hockey sim-to-sim traj}, we show that with lower context differences between the source and target, the point-wise L2 trajectory distance also becomes smaller accordingly. 
\abbv are able to improve trajectory differences by about $40\%$ compared to identification baselines, and $50\%$ compared to DR. Given the parameter estimation error shown in Figure~\ref{fig:push_one_context_align}, the significant trajectory difference is expected from baseline methods.

\textbf{Ablation results.}
In object scooping experiments, we observe that the linear interpolation approach converges more slowly in terms of adaptation iterations, as shown in Figure~\ref{fig:combined_scooping}. 
Due to limited exploration, it hinders performance. 
Figure~\ref{fig:push_one_context_align} shows that \abbv + linear interpolation follows a near-linear sim-to-real transition until iteration 7, closely matching the dataset's transition history. However, it struggles to establish a robust search process due to overfitting to linear interpolated transition histories. 
Except for left damping parameter, no significant performance differences are seen between randomized binary search (ours) and standard binary search. With added randomized, it did not hinder the estimation performance, whereas it learned a more robust adaptation process.

\subsection{Sim-to-Real SysID Evaluation}
\label{Sim-to-Real SysID Evaluation}
We evaluate the task performance during sim-to-real SysID in real-world setups of object scooping and table air hockey. Our method has shown significant performance improvement on trajectory alignment and success rate compared to baseline methods. We evaluated all of our baselines in the sim-to-real transfer.

\begin{table}[tp]
\centering
\caption{Sim-to-Real Object Scooping Success Rate. \textbf{Bold} font means the best.}
\label{tab:real world scoop}
\resizebox{\linewidth}{!}{  
\begin{tabular}{c|c|ccccc}
\toprule
\multirow{2}{*}{\centering Scooping Objects} & \multirow{2}{*}{\centering Method} & \multicolumn{5}{c}{Adaptation iterations} \\ 
 & & 1 iteration & 3 iterations & 5 iterations & 7 iterations & 9 iterations \\ 
\midrule
\multirow{4}{*}{\centering Eggplant} 
                          & \centering ED   & 0.8           & 0.5           & 0.3           & 0.2  & 0.1  \\
                          & \centering DR   & \textbf{0.9} & \textbf{0.9} & \textbf{1.0} & 0.9 & 0.9 \\
                          & \centering TuneNet   & 0.9           & 0.9           & 0.7           & 0.6  & 0.4  \\
                          & \centering \abbv & 0.3           & 0.6           & 0.9           & \textbf{0.9}  & \textbf{0.9}  \\ \hline
\multirow{4}{*}{\centering Celery}   
                          & \centering ED   & 0.0           & 0.1           & 0.3           & 0.1  & 0.2  \\
                          & \centering DR   & 0.0            & 0.2            & 0.0            & 0.1 & 0.0 \\
                          & \centering TuneNet   & 0.0           & 0.1           & 0.3           & 0.1  & 0.1  \\
                          & \centering \abbv & 0.0  & \textbf{0.4}  & \textbf{0.6}  & \textbf{0.7} & \textbf{0.7} \\ \hline
\multirow{4}{*}{\centering Carrot}   
                          & \centering ED   & 0.0           & 0.8           & 0.5           & 0.5  & 0.6  \\
                          & \centering DR   & 0.0            & 0.1            & 0.0            & 0.1  & 0.0  \\
                          & \centering TuneNet   & 0.0           & \textbf{0.9}           & 0.8           & 0.5  & 0.4  \\
                          & \centering \abbv & \textbf{0.3}  & 0.7  & \textbf{0.9}  & \textbf{0.9}  & \textbf{0.9} \\
\bottomrule
\end{tabular}
}
\end{table}

\textbf{Object Scooping Sim-to-Real Evaluation.}
In this experiment, we verify that \abbv can accurately identify the center of mass across various objects during scooping.  
To verify the effectiveness of our algorithm, we selected three different objects (i.e.,celery, carrot, and eggplant) with asymmetrical properties to ensure the difficulty of identifying the center of mass. 
We evaluated each object ten times starting at the absolute center point. 
Similarly to the sim-to-sim transfer setting, we use task success rate to reflect the task performance instead of trajectory matching. 
To obtain the real-world object 3D pose, we utilize a point cloud to localize the object and provide the tilting direction labels.

Inspecting Table~\ref{tab:real world scoop}, we find that DR excels when scooping objects with centralized centers of mass, such as the eggplant, achieving a success rate of 90\% or higher from just 1 iteration.
However, for objects with more complex mass distributions (i.e.,celery and carrot), DR's performance drops significantly.
\abbv is able to adapt to different objects and achieve at least 70\% at 7th iterations.
After successfully lifting the object, one-step adaptation methods randomly sample other parameter values due to the absence of history and lack of target-to-target parameter transition during training, while \abbv consistently lifts the object in subsequent iterations. 
This performance demonstrates \abbv’s ability to generalize to unseen scenarios (target-to-target adaptation) by leveraging context history. 
Its ability to maintain high success rates, especially with objects that have complex mass properties, underscores its effectiveness in real-world scooping tasks.

\textbf{Air Hockey Sim-to-Real Evaluation.}
We set up the real-world table air hockey as shown in Figure~\ref{fig:combined_env}. To create varying friction across the two surfaces, we installed separate fans under each side of the table, with adjustable fan voltages controlling the sliding friction. 
We evaluated the sim-to-real transfer performance over 15 trials using 3 different seeds, with each trial having randomized fan voltages on both sides.
\begin{table}[tp]
\centering
\caption{Sim-to-Real Table Air Hockey Trajectory Differences in Meters over 15 Runs. \textbf{Bold} font means the best.}
\label{tab:real world airhockey exp}
\resizebox{\linewidth}{!}{  
\begin{tabular}{c|ccccc}
\toprule
\multicolumn{1}{c|}{\multirow{2}{*}{Method}} & \multicolumn{5}{c}{Adaptation iterations}                                                                                                 \\  
\multicolumn{1}{l|}{}                        & 1 iteration                   & 3 iterations                  & 5 iterations                  & 7 iterations                  & 9 iterations                  \\ 
\midrule
ED                                           & \textbf{0.40 $\pm$ 0.19} & \textbf{0.34 $\pm$ 0.14} & 0.51 $\pm$ 0.44          & 0.34 $\pm$ 0.15          & 0.34 $\pm$ 0.14          \\ \hline
DR                                           & 0.41 $\pm$ 0.27            & 0.40 $\pm$ 0.1            & 0.37 $\pm$ 0.33            & 0.42 $\pm$ 0.42            & 0.43 $\pm$ 0.40            \\ \hline
TuneNet                                           & 0.47 $\pm$ 0.22            & 0.40 $\pm$ 0.11            & \textbf{0.32 $\pm$ 0.21}            & 0.38 $\pm$ 0.16            & 0.34 $\pm$ 0.15            \\ \hline
\abbv                         & 0.47 $\pm$ 0.18          & 0.35 $\pm$ 0.14          & 0.35 $\pm$ 0.12 & \textbf{0.29 $\pm$ 0.10} & \textbf{0.27 $\pm$ 0.10}\\
\bottomrule
\end{tabular}
}
\end{table}
The results from the sim-to-real air hockey experiment, presented in Table~\ref{tab:real world airhockey exp}, show the performance of different methods in trajectory matching over multiple adaptation iterations. 
For one-iteration adaptation, ED performs best with a trajectory difference of 0.40, as it tries to adapts to the target parameter within one iteration. However, as iterations increase, \abbv steadily improves, outperforming the baselines. By the 7th and 9th iterations, \abbv achieves the lowest trajectory differences of 0.29 and 0.27, respectively. In the final iterations, \abbv delivers about 20\% better performance than the top baseline methods.
\section{Conclusion}
This paper introduces a novel in-context learning approach to bridge the sim-to-real gap in robotic tasks by adjusting environment parameters online. By leveraging interaction histories as context, we enable dynamics adaptation to real-world environments without requiring model updates. 
Evaluated in scooping and table air-hockey tasks, our method outperforms traditional approaches such as domain randomization and TuneNet, reducing the sim-to-real gap and improving both sim-to-sim and sim-to-real performance. 
The approach leverages historical multi-episode data to infer system parameters and provide a better real-world dynamics prediction. 
While our method demonstrates strong performance, further alignment is required between state trajectories and environment parameter spaces to extend the approach to multi-task settings, enabling one SysID model to handle multiple task environments. Nonetheless, the framework provides a more efficient and accurate solution for the real-world deployment of simulation-based robotic systems.\looseness=-1 
\section{Acknowledgement}

The authors want to acknowledge the support from the National Science Foundation under grants CNS-2047454. We also thank Haohong Lin and Changyi Lin for discussion.

\bibliography{reference} 

\begin{thebibliography}{10}
\providecommand{\url}[1]{#1}
\csname url@samestyle\endcsname
\providecommand{\newblock}{\relax}
\providecommand{\bibinfo}[2]{#2}
\providecommand{\BIBentrySTDinterwordspacing}{\spaceskip=0pt\relax}
\providecommand{\BIBentryALTinterwordstretchfactor}{4}
\providecommand{\BIBentryALTinterwordspacing}{\spaceskip=\fontdimen2\font plus
\BIBentryALTinterwordstretchfactor\fontdimen3\font minus \fontdimen4\font\relax}
\providecommand{\BIBforeignlanguage}[2]{{%
\expandafter\ifx\csname l@#1\endcsname\relax
\typeout{** WARNING: IEEEtran.bst: No hyphenation pattern has been}%
\typeout{** loaded for the language `#1'. Using the pattern for}%
\typeout{** the default language instead.}%
\else
\language=\csname l@#1\endcsname
\fi
#2}}
\providecommand{\BIBdecl}{\relax}
\BIBdecl

\bibitem{8460528}
X.~B. Peng, M.~Andrychowicz, W.~Zaremba, and P.~Abbeel, ``Sim-to-real transfer of robotic control with dynamics randomization,'' in \emph{2018 IEEE International Conference on Robotics and Automation (ICRA)}, 2018, pp. 3803--3810.

\bibitem{lin2024generalize}
H.~Lin, R.~Corcodel, and D.~Zhao, ``Generalize by touching: Tactile ensemble skill transfer for robotic furniture assembly,'' \emph{arXiv preprint arXiv:2404.17684}, 2024.

\bibitem{10611442}
Y.~Li, J.~Li, W.~Fu, and Y.~Wu, ``Learning agile bipedal motions on a quadrupedal robot,'' in \emph{2024 IEEE International Conference on Robotics and Automation (ICRA)}, 2024, pp. 9735--9742.

\bibitem{kumar2021rma}
A.~Kumar, Z.~Fu, D.~Pathak, and J.~Malik, ``Rma: Rapid motor adaptation for legged robots,'' \emph{arXiv preprint arXiv:2107.04034}, 2021.

\bibitem{chen2024learning}
Z.~Chen, X.~He, Y.-J. Wang, Q.~Liao, Y.~Ze, Z.~Li, S.~S. Sastry, J.~Wu, K.~Sreenath, S.~Gupta \emph{et~al.}, ``Learning smooth humanoid locomotion through lipschitz-constrained policies,'' \emph{arXiv preprint arXiv:2410.11825}, 2024.

\bibitem{zhang2024wococo}
C.~Zhang, W.~Xiao, T.~He, and G.~Shi, ``Wococo: Learning whole-body humanoid control with sequential contacts,'' \emph{arXiv preprint arXiv:2406.06005}, 2024.

\bibitem{xu2022trustworthy}
M.~Xu, Z.~Liu, P.~Huang, W.~Ding, Z.~Cen, B.~Li, and D.~Zhao, ``Trustworthy reinforcement learning against intrinsic vulnerabilities: Robustness, safety, and generalizability,'' \emph{arXiv preprint arXiv:2209.08025}, 2022.

\bibitem{wang2023guardians}
Y.~Wang, M.~Xu, G.~Shi, and D.~Zhao, ``Guardians as you fall: Active mode transition for safe falling,'' \emph{arXiv preprint arXiv:2310.04828}, 2023.

\bibitem{yao2024constraint}
Y.~Yao, Z.~Liu, Z.~Cen, J.~Zhu, W.~Yu, T.~Zhang, and D.~Zhao, ``Constraint-conditioned policy optimization for versatile safe reinforcement learning,'' \emph{Advances in Neural Information Processing Systems}, vol.~36, 2024.

\bibitem{hu2024learning}
H.~Hu, X.~Zhang, X.~Lyu, and M.~Chen, ``Learning robust policies via interpretable hamilton-jacobi reachability-guided disturbances,'' \emph{arXiv preprint arXiv:2409.19746}, 2024.

\bibitem{torne2024reconciling}
M.~Torne, A.~Simeonov, Z.~Li, A.~Chan, T.~Chen, A.~Gupta, and P.~Agrawal, ``Reconciling reality through simulation: A real-to-sim-to-real approach for robust manipulation,'' \emph{arXiv preprint arXiv:2403.03949}, 2024.

\bibitem{huang2022robust}
P.~Huang, M.~Xu, F.~Fang, and D.~Zhao, ``Robust reinforcement learning as a stackelberg game via adaptively-regularized adversarial training,'' in \emph{the 31st International Joint Conference on Artificial Intelligence (IJCAI)}.\hskip 1em plus 0.5em minus 0.4em\relax Proceedings of the Thirty-First International Joint Conference on Artificial~…, 2022.

\bibitem{pmlr-v100-mehta20a}
\BIBentryALTinterwordspacing
B.~Mehta, M.~Diaz, F.~Golemo, C.~J. Pal, and L.~Paull, ``Active domain randomization,'' in \emph{Proceedings of the Conference on Robot Learning}, ser. Proceedings of Machine Learning Research, L.~P. Kaelbling, D.~Kragic, and K.~Sugiura, Eds., vol. 100.\hskip 1em plus 0.5em minus 0.4em\relax PMLR, 30 Oct--01 Nov 2020, pp. 1162--1176. [Online]. Available: \url{https://proceedings.mlr.press/v100/mehta20a.html}
\BIBentrySTDinterwordspacing

\bibitem{ramos2019bayessim}
F.~Ramos, R.~C. Possas, and D.~Fox, ``Bayessim: adaptive domain randomization via probabilistic inference for robotics simulators,'' \emph{arXiv preprint arXiv:1906.01728}, 2019.

\bibitem{pmlr-v229-huang23c}
\BIBentryALTinterwordspacing
P.~Huang, X.~Zhang, Z.~Cao, S.~Liu, M.~Xu, W.~Ding, J.~Francis, B.~Chen, and D.~Zhao, ``What went wrong? closing the sim-to-real gap via differentiable causal discovery,'' in \emph{Proceedings of The 7th Conference on Robot Learning}, ser. Proceedings of Machine Learning Research, J.~Tan, M.~Toussaint, and K.~Darvish, Eds., vol. 229.\hskip 1em plus 0.5em minus 0.4em\relax PMLR, 06--09 Nov 2023, pp. 734--760. [Online]. Available: \url{https://proceedings.mlr.press/v229/huang23c.html}
\BIBentrySTDinterwordspacing

\bibitem{dong2022survey}
Q.~Dong, L.~Li, D.~Dai, C.~Zheng, Z.~Wu, B.~Chang, X.~Sun, J.~Xu, and Z.~Sui, ``A survey on in-context learning,'' \emph{arXiv preprint arXiv:2301.00234}, 2022.

\bibitem{laskin2022context}
M.~Laskin, L.~Wang, J.~Oh, E.~Parisotto, S.~Spencer, R.~Steigerwald, D.~Strouse, S.~Hansen, A.~Filos, E.~Brooks \emph{et~al.}, ``In-context reinforcement learning with algorithm distillation,'' \emph{arXiv preprint arXiv:2210.14215}, 2022.

\bibitem{grigsbyamago}
J.~Grigsby, L.~Fan, and Y.~Zhu, ``Amago: Scalable in-context reinforcement learning for adaptive agents,'' in \emph{The Twelfth International Conference on Learning Representations}.

\bibitem{xu2022prompting}
M.~Xu, Y.~Shen, S.~Zhang, Y.~Lu, D.~Zhao, J.~Tenenbaum, and C.~Gan, ``Prompting decision transformer for few-shot policy generalization,'' in \emph{international conference on machine learning}.\hskip 1em plus 0.5em minus 0.4em\relax PMLR, 2022, pp. 24\,631--24\,645.

\bibitem{chen2021decision}
L.~Chen, K.~Lu, A.~Rajeswaran, K.~Lee, A.~Grover, M.~Laskin, P.~Abbeel, A.~Srinivas, and I.~Mordatch, ``Decision transformer: Reinforcement learning via sequence modeling,'' \emph{Advances in neural information processing systems}, vol.~34, pp. 15\,084--15\,097, 2021.

\bibitem{fu2024context}
L.~Fu, H.~Huang, G.~Datta, L.~Y. Chen, W.~C.-H. Panitch, F.~Liu, H.~Li, and K.~Goldberg, ``In-context imitation learning via next-token prediction,'' \emph{arXiv preprint arXiv:2408.15980}, 2024.

\bibitem{kumar2022adapting}
A.~Kumar, Z.~Li, J.~Zeng, D.~Pathak, K.~Sreenath, and J.~Malik, ``Adapting rapid motor adaptation for bipedal robots,'' in \emph{2022 IEEE/RSJ International Conference on Intelligent Robots and Systems (IROS)}.\hskip 1em plus 0.5em minus 0.4em\relax IEEE, 2022, pp. 1161--1168.

\bibitem{pmlr-v229-ren23b}
\BIBentryALTinterwordspacing
A.~Z. Ren, H.~Dai, B.~Burchfiel, and A.~Majumdar, ``Adaptsim: Task-driven simulation adaptation for sim-to-real transfer,'' in \emph{Proceedings of The 7th Conference on Robot Learning}, ser. Proceedings of Machine Learning Research, J.~Tan, M.~Toussaint, and K.~Darvish, Eds., vol. 229.\hskip 1em plus 0.5em minus 0.4em\relax PMLR, 06--09 Nov 2023, pp. 3434--3452. [Online]. Available: \url{https://proceedings.mlr.press/v229/ren23b.html}
\BIBentrySTDinterwordspacing

\bibitem{murooka2021exi}
T.~Murooka, M.~Hamaya, F.~von Drigalski, K.~Tanaka, and Y.~Ijiri, ``Exi-net: Explicitly/implicitly conditioned network for multiple environment sim-to-real transfer,'' in \emph{Conference on Robot Learning}.\hskip 1em plus 0.5em minus 0.4em\relax PMLR, 2021, pp. 1221--1230.

\bibitem{chebotar2019closing}
Y.~Chebotar, A.~Handa, V.~Makoviychuk, M.~Macklin, J.~Issac, N.~Ratliff, and D.~Fox, ``Closing the sim-to-real loop: Adapting simulation randomization with real world experience,'' in \emph{2019 International Conference on Robotics and Automation (ICRA)}.\hskip 1em plus 0.5em minus 0.4em\relax IEEE, 2019, pp. 8973--8979.

\bibitem{muratore2022neural}
F.~Muratore, T.~Gruner, F.~Wiese, B.~Belousov, M.~Gienger, and J.~Peters, ``Neural posterior domain randomization,'' in \emph{Conference on Robot Learning}.\hskip 1em plus 0.5em minus 0.4em\relax PMLR, 2022, pp. 1532--1542.

\bibitem{lim2022real2sim2real}
V.~Lim, H.~Huang, L.~Y. Chen, J.~Wang, J.~Ichnowski, D.~Seita, M.~Laskey, and K.~Goldberg, ``Real2sim2real: Self-supervised learning of physical single-step dynamic actions for planar robot casting,'' in \emph{2022 International Conference on Robotics and Automation (ICRA)}.\hskip 1em plus 0.5em minus 0.4em\relax IEEE, 2022, pp. 8282--8289.

\bibitem{yu2017preparing}
W.~Yu, J.~Tan, C.~K. Liu, and G.~Turk, ``Preparing for the unknown: Learning a universal policy with online system identification,'' \emph{arXiv preprint arXiv:1702.02453}, 2017.

\bibitem{evans2022context}
B.~Evans, A.~Thankaraj, and L.~Pinto, ``Context is everything: Implicit identification for dynamics adaptation,'' in \emph{2022 International Conference on Robotics and Automation (ICRA)}.\hskip 1em plus 0.5em minus 0.4em\relax IEEE, 2022, pp. 2642--2648.

\bibitem{allevato2020tunenet}
A.~Allevato, E.~S. Short, M.~Pryor, and A.~Thomaz, ``Tunenet: One-shot residual tuning for system identification and sim-to-real robot task transfer,'' in \emph{Conference on Robot Learning}.\hskip 1em plus 0.5em minus 0.4em\relax PMLR, 2020, pp. 445--455.

\bibitem{memmelasid}
M.~Memmel, A.~Wagenmaker, C.~Zhu, D.~Fox, and A.~Gupta, ``Asid: Active exploration for system identification in robotic manipulation,'' in \emph{The Twelfth International Conference on Learning Representations}.

\bibitem{jiang2024transic}
Y.~Jiang, C.~Wang, R.~Zhang, J.~Wu, and L.~Fei-Fei, ``Transic: Sim-to-real policy transfer by learning from online correction,'' \emph{arXiv preprint arXiv:2405.10315}, 2024.

\bibitem{dai2024acdc}
T.~Dai, J.~Wong, Y.~Jiang, C.~Wang, C.~Gokmen, R.~Zhang, J.~Wu, and L.~Fei-Fei, ``Acdc: Automated creation of digital cousins for robust policy learning,'' \emph{arXiv preprint arXiv:2410.07408}, 2024.

\bibitem{krishnamurthy2024can}
A.~Krishnamurthy, K.~Harris, D.~J. Foster, C.~Zhang, and A.~Slivkins, ``Can large language models explore in-context?'' \emph{arXiv preprint arXiv:2403.15371}, 2024.

\bibitem{wang2023images}
X.~Wang, W.~Wang, Y.~Cao, C.~Shen, and T.~Huang, ``Images speak in images: A generalist painter for in-context visual learning,'' in \emph{Proceedings of the IEEE/CVF Conference on Computer Vision and Pattern Recognition}, 2023, pp. 6830--6839.

\bibitem{min2022rethinking}
S.~Min, X.~Lyu, A.~Holtzman, M.~Artetxe, M.~Lewis, H.~Hajishirzi, and L.~Zettlemoyer, ``Rethinking the role of demonstrations: What makes in-context learning work?'' \emph{arXiv preprint arXiv:2202.12837}, 2022.

\bibitem{xu2023hyper}
M.~Xu, Y.~Lu, Y.~Shen, S.~Zhang, D.~Zhao, and C.~Gan, ``Hyper-decision transformer for efficient online policy adaptation,'' \emph{arXiv preprint arXiv:2304.08487}, 2023.

\bibitem{zhu2024incoro}
J.~Y. Zhu, C.~G. Cano, D.~V. Bermudez, and M.~Drozdzal, ``Incoro: In-context learning for robotics control with feedback loops,'' \emph{arXiv preprint arXiv:2402.05188}, 2024.

\bibitem{di2024keypoint}
N.~Di~Palo and E.~Johns, ``Keypoint action tokens enable in-context imitation learning in robotics,'' \emph{arXiv preprint arXiv:2403.19578}, 2024.

\bibitem{yu2024fewshotincontextpreferencelearning}
\BIBentryALTinterwordspacing
C.~Yu, H.~Lu, J.~Gao, Q.~Tan, X.~Yang, Y.~Wang, Y.~Wu, and E.~Vinitsky, ``Few-shot in-context preference learning using large language models,'' 2024. [Online]. Available: \url{https://arxiv.org/abs/2410.17233}
\BIBentrySTDinterwordspacing

\bibitem{jiang2023learning}
C.~Jiang, N.~R. Ke, and H.~van Hasselt, ``Learning how to infer partial mdps for in-context adaptation and exploration,'' \emph{arXiv preprint arXiv:2302.04250}, 2023.

\bibitem{zhu2020robosuite}
Y.~Zhu, J.~Wong, A.~Mandlekar, R.~Mart{\'\i}n-Mart{\'\i}n, A.~Joshi, S.~Nasiriany, and Y.~Zhu, ``robosuite: A modular simulation framework and benchmark for robot learning,'' \emph{arXiv preprint arXiv:2009.12293}, 2020.

\bibitem{haarnoja2018soft}
T.~Haarnoja, A.~Zhou, P.~Abbeel, and S.~Levine, ``Soft actor-critic: Off-policy maximum entropy deep reinforcement learning with a stochastic actor,'' in \emph{International conference on machine learning}.\hskip 1em plus 0.5em minus 0.4em\relax PMLR, 2018, pp. 1861--1870.

\bibitem{martinez1998randomized}
C.~Mart{\'\i}nez and S.~Roura, ``Randomized binary search trees,'' \emph{Journal of the ACM (JACM)}, vol.~45, no.~2, pp. 288--323, 1998.

\bibitem{radford2019language}
A.~Radford, J.~Wu, R.~Child, D.~Luan, D.~Amodei, I.~Sutskever \emph{et~al.}, ``Language models are unsupervised multitask learners,'' \emph{OpenAI blog}, vol.~1, no.~8, p.~9, 2019.

\bibitem{al2019character}
R.~Al-Rfou, D.~Choe, N.~Constant, M.~Guo, and L.~Jones, ``Character-level language modeling with deeper self-attention,'' in \emph{Proceedings of the AAAI conference on artificial intelligence}, vol.~33, no.~01, 2019, pp. 3159--3166.

\bibitem{shi2023robocook}
H.~Shi, H.~Xu, S.~Clarke, Y.~Li, and J.~Wu, ``Robocook: Long-horizon elasto-plastic object manipulation with diverse tools,'' in \emph{Conference on Robot Learning}.\hskip 1em plus 0.5em minus 0.4em\relax PMLR, 2023, pp. 642--660.

\bibitem{chuck2024robot}
C.~Chuck, C.~Qi, M.~J. Munje, S.~Li, M.~Rudolph, C.~Shi, S.~Agarwal, H.~Sikchi, A.~Peri, S.~Dayal \emph{et~al.}, ``Robot air hockey: A manipulation testbed for robot learning with reinforcement learning,'' \emph{arXiv preprint arXiv:2405.03113}, 2024.

\bibitem{sikorski1982bisection}
K.~Sikorski, ``Bisection is optimal,'' \emph{Numerische Mathematik}, vol.~40, pp. 111--117, 1982.

\end{thebibliography}
\clearpage

\section{Appendix}

\subsection{Analysis on Different Transition Sequence Length $L$}
\begin{figure*}[]
    \centering
    \includegraphics[width=\linewidth]{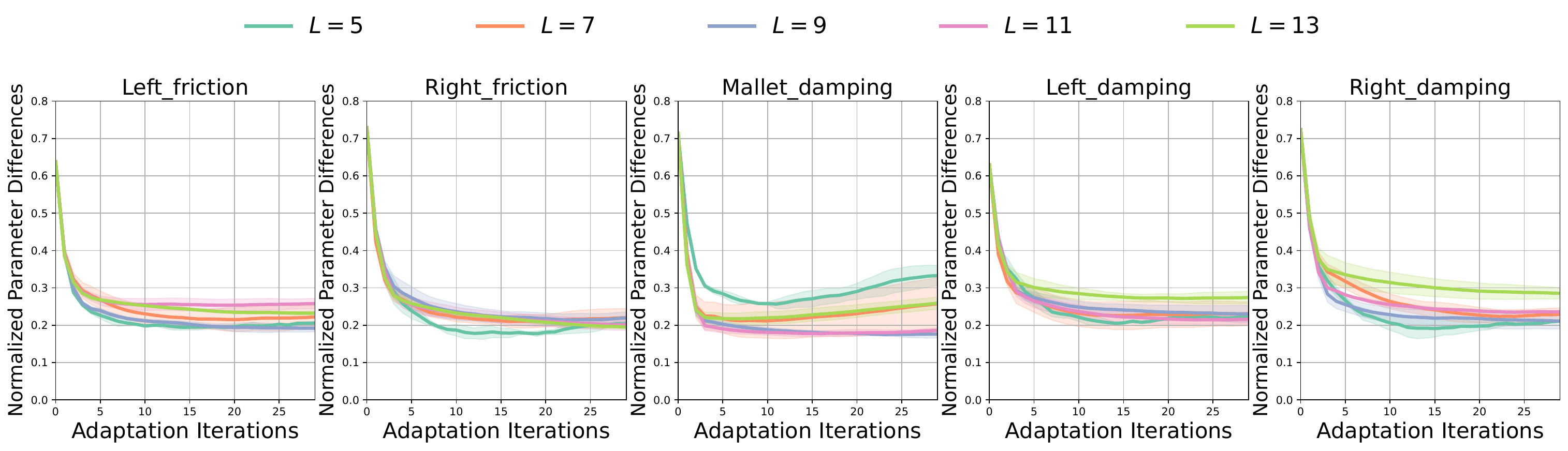}
    \caption{Table air hockey sim-to-sim transfer SysID performance across different parameter transition sequence lengths $L$.}
    \label{fig:figs/appendix_image_different_K_push_one}
\end{figure*}

We conducted ablation studies on the transition sequence length $L$ to validate the chosen hyperparameters in our experiments. The sequence length $L$ was varied from 5 to 13, as shown in Figure~\ref{fig:figs/appendix_image_different_K_push_one}. We collected 100 pairs of simulation and simulated real environment parameters to comprehensively evaluate performance under different initial conditions. The results demonstrate minimal performance variation across different values of $L$, with some degradation observed at $L = 5$ and $L = 13$. Transition sequence lengths between 7 and 9 consistently yielded stable performance.

This range aligns well with other settings, as the proposed randomized binary search algorithm is independent of the dimensionality of state trajectories, action spaces, and environment parameter spaces.

A similar pattern was observed in the object scooping task, shown in Figure~\ref{fig:figs/appendix_image_different_K_scoop}. Parameter estimation performance followed similar trends, further validating the choice of transition sequence length.

\subsection{Analysis on Different Transformer Window Sizes $k$}

\begin{figure*}[ht]
    \centering
    \includegraphics[width=\linewidth]{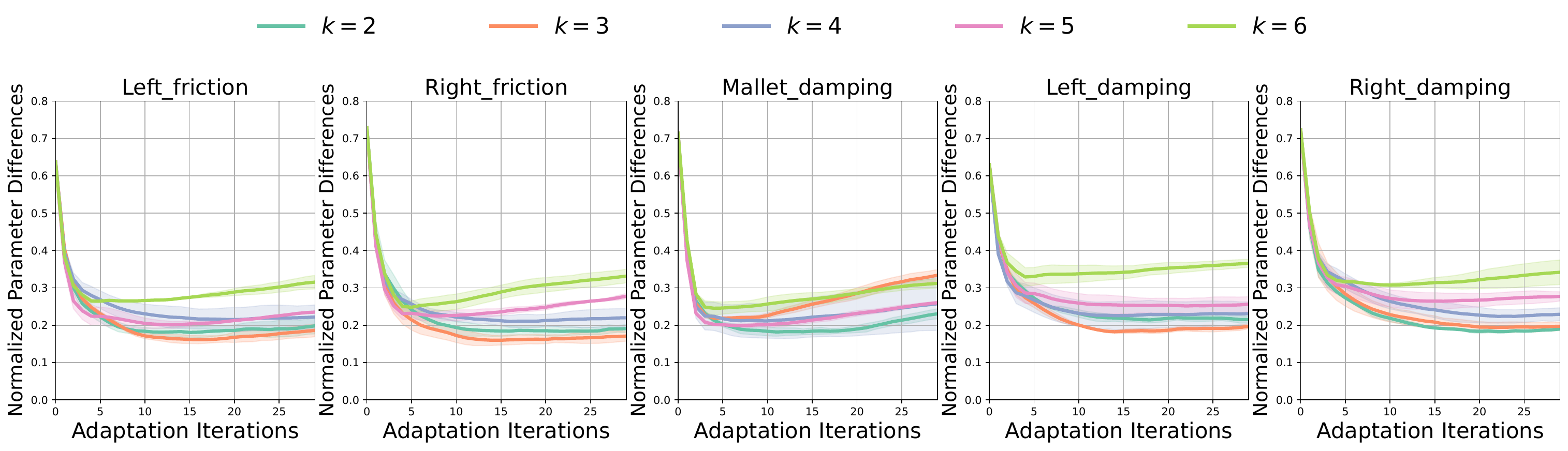}
    \caption{Table air hockey sim-to-sim transfer SysID performance across different window sizes.}
    \label{fig:figs/appendix_image_different_k_push}
\end{figure*}

We collected 100 pairs of sim and simulated real environment parameters to fully assess performance under different initial conditions. To evaluate the performance dependency on the length of the window size $k$, we kept the generated transition sequence length $L = 7$ and varied the window size $k$ from 2 to 6, as shown in Figures~\ref{fig:figs/appendix_image_different_k_push} and Figure~\ref{fig:figs/appendix_image_different_k_scoop}. We can observe that ($k = 2$) and ($k = 3$) perform better on most of the environment parameters, whereas the longer window size tends to perform worse, especially in later iterations. This is a reasonable behavior because larger window sizes are easier to overfit to the training data sequence (in this case, $L=7$) and lead to worse performance on Out-of-Distribution(OOD) generalization after adaptation iteration exceeds $L$. Conversely, a smaller window size learns from a sampled subsequence from the whole generated transition sequences, which focuses on local features rather than long-horizon features. Such structure would enhance the generalizability to OOD settings. Conclusively, the value of window size depends on the state trajectory complexity of the dynamical systems. If the environment requires more steps to explore the environments, it would generally need more history(window size) to predict the next-step environment parameters and vice versa. 

\subsection{Comparison with State-of-the-Art Offline SysID Baselines}
\begin{figure*}[]
    \centering
    \includegraphics[width=\linewidth]{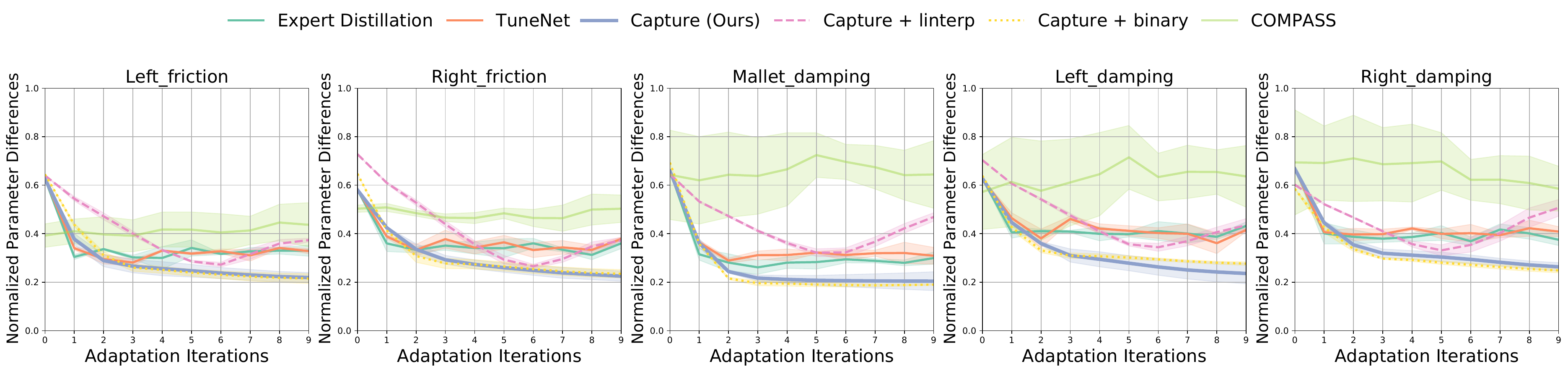}
    \caption{Table air hockey sim-to-sim transfer with added baseline.}
    \label{fig:figs/appendix_image_push_one_compass}
\end{figure*}

We included COMPASS~\cite{pmlr-v229-huang23c} as an extra baseline for comparison. In the original paper, the authors proposed to collect 10 trajectories to estimate the environment parameters. With the CAPTURE setting, since we only need to collect one trajectory at every iteration, we will compare the results with COMPASS through 10 iteration adaptation sequences. Moreover, since we only provided the ``useful" environment parameters to CAPTURE, we also set the COMPASS causal graph as full\_graph to ensure that all environment parameters will be optimized. We evaluated COMPASS with 50 random source and target environment parameters across 3 different seeds. The results for both experiments are shown in Figures~\ref{fig:figs/appendix_image_push_one_compass} and Figure~\ref{fig:figs/appendix_image_scoop_compass}. We can observe that COMPASS had a good trend of aligning the center of mass in the object scooping tasks but converging comparatively slower. In the table air hockey setting, COMPASS was not able to perform well in this setting even though some of the environment parameters have good starting points (left\_friction). 

\subsection{Limitations of Monotonicity Assumptions Between Environment Parameters and State Trajectories}

Indeed, the binary search method relies on the monotonic relationship between environment parameters and state trajectories under the same action. We also observed that most physical properties defined in simulators inherently follow such monotonic relationships. For example, in our experiments, parameters like friction and damping consistently influence the behavior of state trajectories in predictable ways. Increasing the damping coefficient typically results in slower system responses, while higher friction values lead to reduced sliding distances. 

While this assumption holds true for many environment parameters, we acknowledge that there may be cases where the relationship between parameters and state trajectories is non-monotonic. Without such an assumption, the binary search process would lack the framework to accurately update these bounds. However, state trajectories are not strictly required when generating transition sequences for the environment parameter identification process, as the ground truth of the environment parameters is accessible. As a result, it is possible to generate reasonable transition sequences even in the presence of a non-monotonic relationship between environment parameters and state trajectories. 

To validate this assumption, we modified the observation of the scoop environment to represent the distance between the scoop point and the center of mass.

As shown in Figure~\ref{fig:figs/appendix_image_original_scoop_dist}, our observations indicate that CAPTURE struggles to identify the correct environment parameter during the first few iterations. However, after the 10th iteration, the error gradually decreases, and CAPTURE still outperforms the baseline methods. Therefore, even in environments with no monotonic relationship between state trajectories and environment parameters, CAPTURE maintains superior performance compared to the baseline approaches.

\subsection{Additional Evaluation Under Noisy Observations}
\begin{figure*}[]
    \centering
    \includegraphics[width=\linewidth]{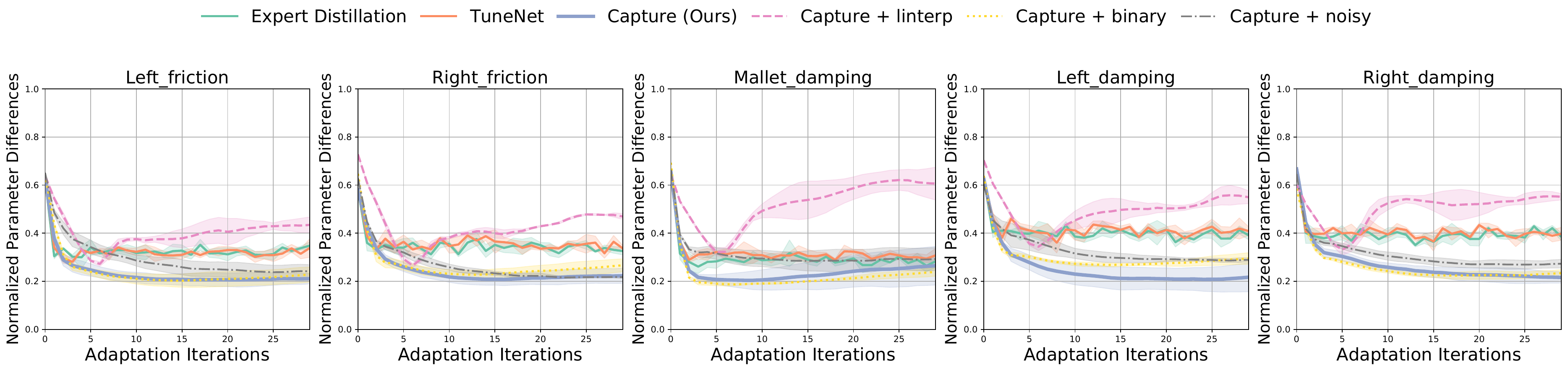}
    \caption{Table air hockey sim-to-sim transfer under noisy observations. }
    \label{fig:figs/appendix_image_push_one_noise}
\end{figure*}

\begin{figure*}[ht]
    \centering
    \begin{subfigure}[b]{0.195\textwidth}
        \centering
        \includegraphics[width=\textwidth]{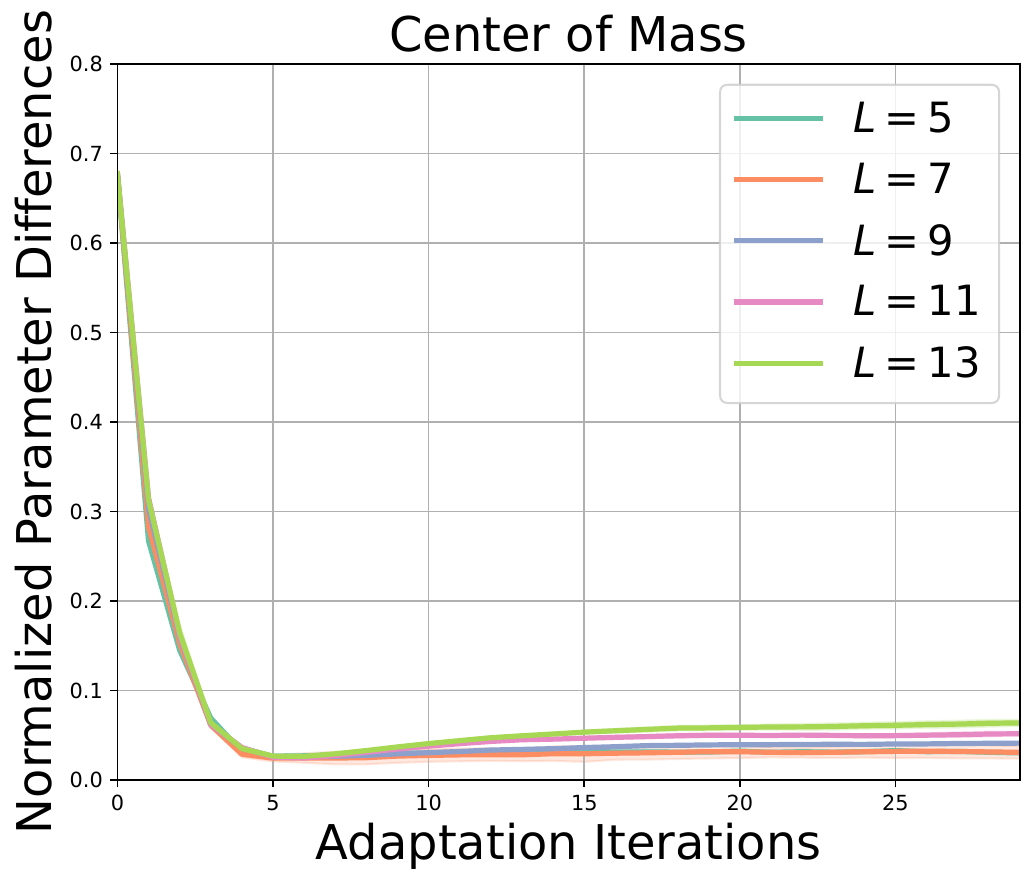}
        \caption{}
        \label{fig:figs/appendix_image_different_K_scoop}
    \end{subfigure}
    \hfill
    \begin{subfigure}[b]{0.195\textwidth}
        \centering
        \includegraphics[width=\textwidth]{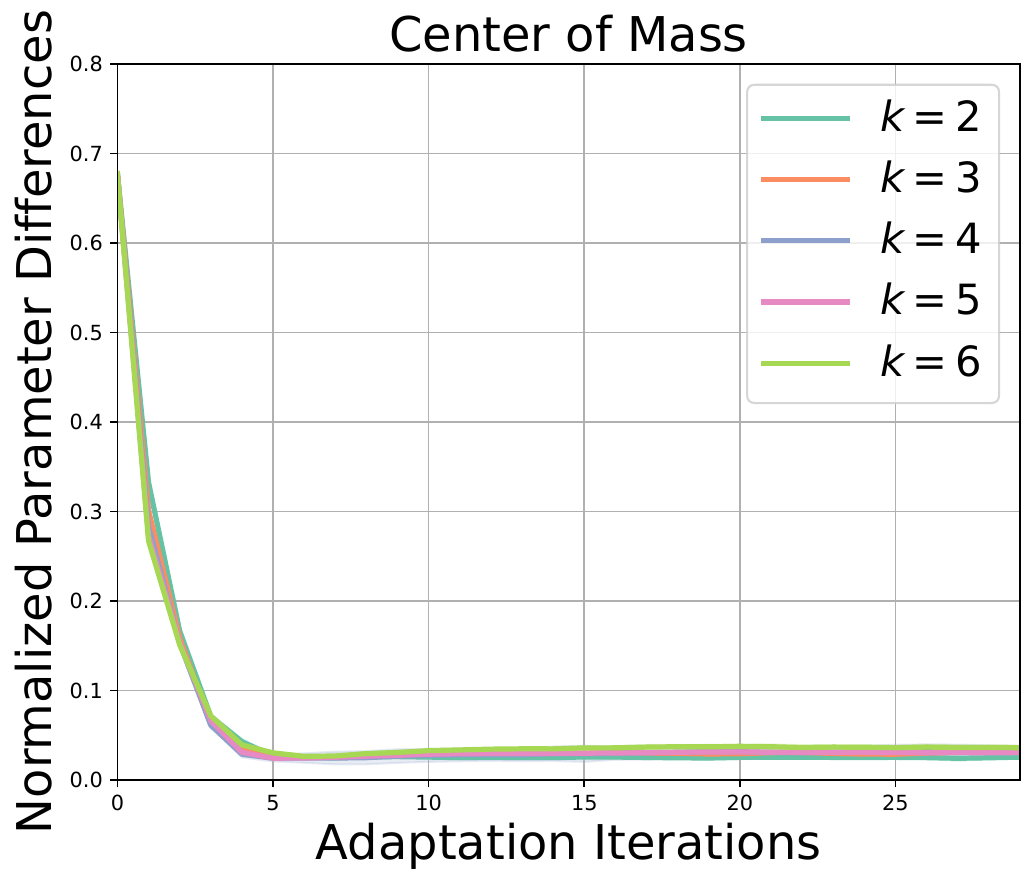}
        \caption{}
        \label{fig:figs/appendix_image_different_k_scoop}
    \end{subfigure}
    \hfill
    \begin{subfigure}[b]{0.195\textwidth}
        \centering
        \includegraphics[width=\textwidth]{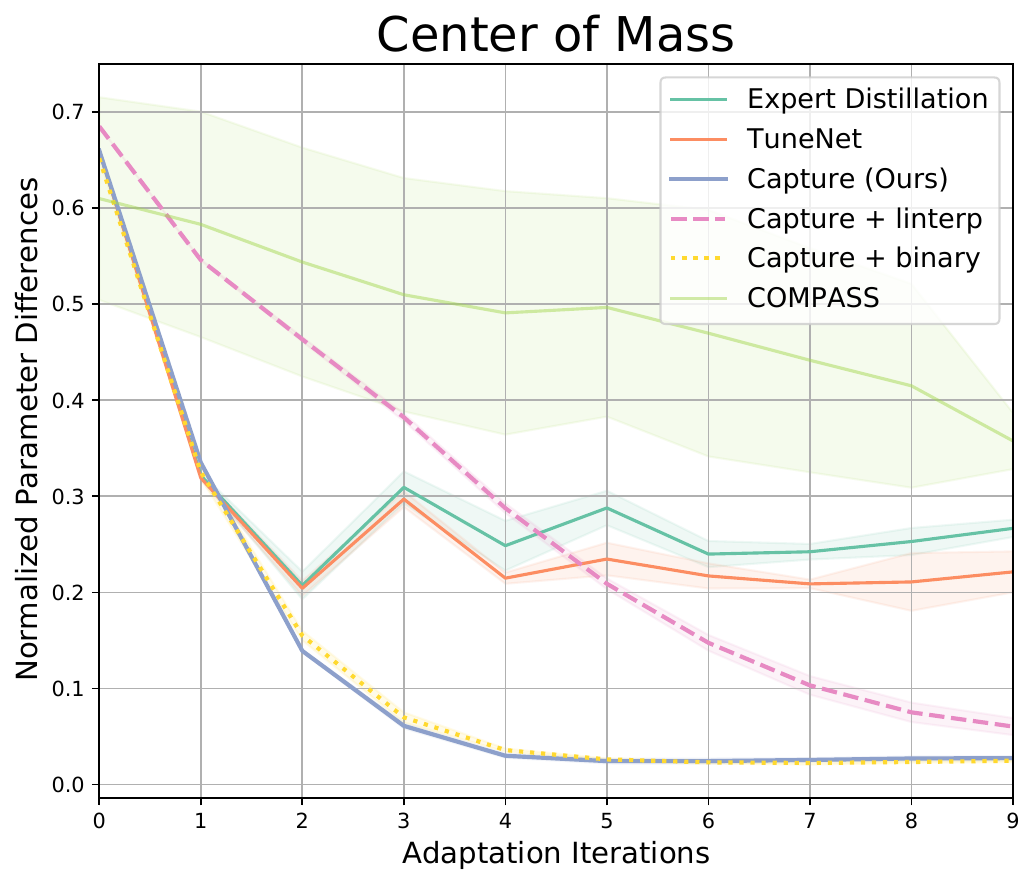}
        \caption{}
        \label{fig:figs/appendix_image_scoop_compass}
    \end{subfigure}
    \hfill
    \begin{subfigure}[b]{0.195\textwidth}
        \centering
        \includegraphics[width=\textwidth]{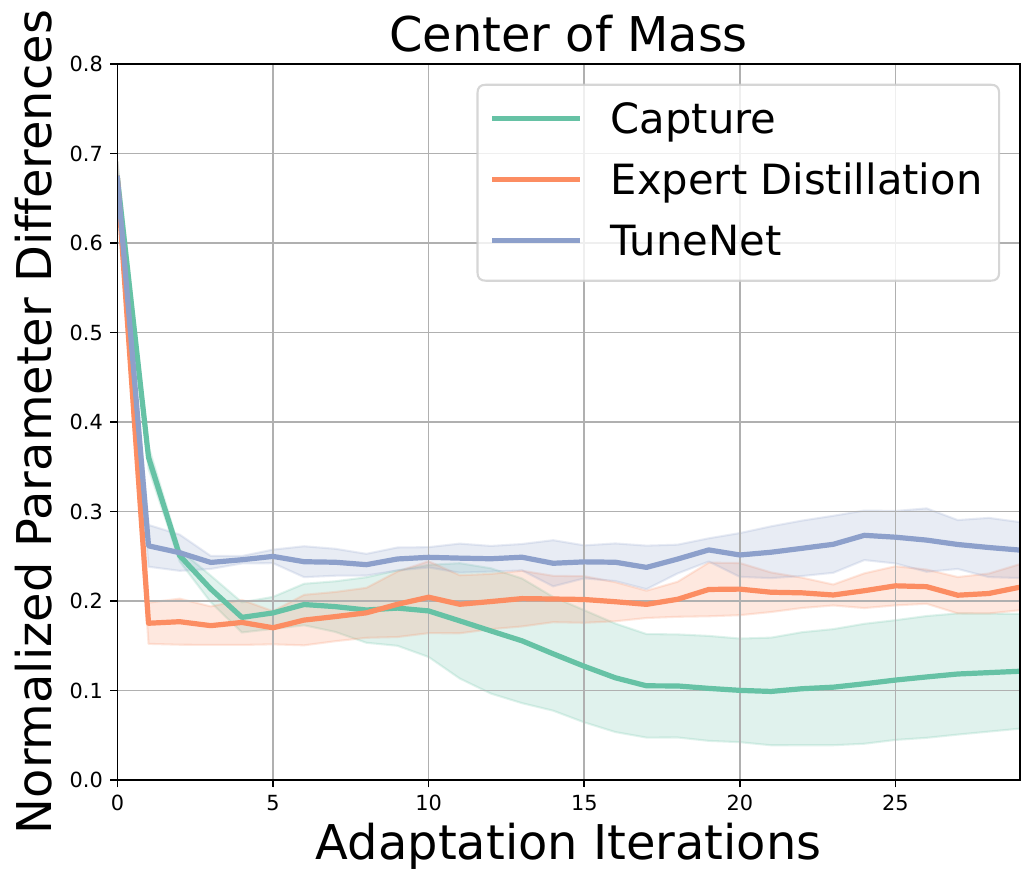}
        \caption{}
        \label{fig:figs/appendix_image_original_scoop_dist}
    \end{subfigure}
    \hfill
    \begin{subfigure}[b]{0.195\textwidth}
        \centering
        \includegraphics[width=\textwidth]{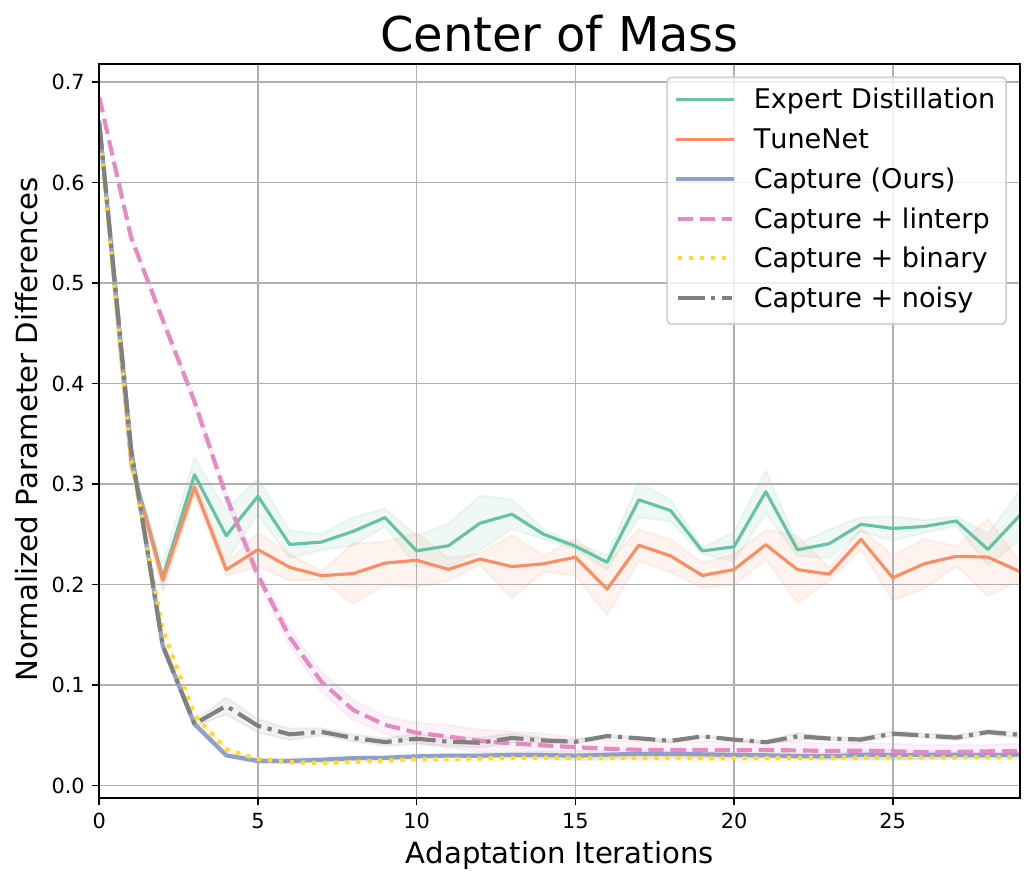}
        \caption{}
        \label{fig:figs/appendix_image_scoop_noise}
    \end{subfigure}
    \caption{
    Object scooping sim-to-sim transfer SysID performance under different variants:
    (a) Performance across varying parameter transition sequence lengths $L$.
    (b) Performance across different window sizes $k$.
    (c) Performance with an added baseline.
    (d) Comparison of CAPTURE, Expert Distillation, and TuneNet in a non-monotonic scooping environment.
    (e) Performance under noisy observation conditions.
    }
\end{figure*}

To further assess the robustness of CAPTURE, we explicitly evaluated the sim-to-sim transfer performance with added noise/perturbation. 
    \begin{itemize}
        \item \textbf{Pose disturbance for object scooping:} Throughout our 30 adaptation iterations, we introduced a random angle label (from 0, -1, 1) every 3 steps to simulate disturbances. At each evaluation (30 adaptation iterations), 10 random labels were generated to act as noise for the angle classifier in the noisy setting. The results are evaluated in the sim-to-sim transfer with 100 randomly generated source-and-target pairs, shown in Figure~\ref{fig:figs/appendix_image_scoop_noise}, reveal slight fluctuations in performance across adaptation iterations. Despite the high frequency of observation failures, CAPTURE consistently maintained robust performance and even outperformed baseline methods that utilized ground-truth state observations.

        \item \textbf{Sensor noise for table air hockey:} In the table air hockey setting, randomly selecting state trajectories would not be a logical approach. Instead, we introduced uniform noise at each iteration to the real state trajectory component, resulting in observed real trajectory values ranging between 90\% and 110\% of the ground-truth values. As shown in Figures~\ref{fig:figs/appendix_image_push_one_noise}, this significant noise slows down the adaptation process and slightly impacts performance. However, similar to the object scooping scenario, CAPTURE demonstrates resilience to noisy observations. By the later adaptation iterations, CAPTURE outperforms the baseline methods by a substantial margin, highlighting its robustness in noisy environments.

    \end{itemize}

\end{document}